\documentclass{article} 
\usepackage{arxiv,times}

\usepackage{amsmath,amsfonts,bm}

\def\eqref#1{equation~\ref{#1}}

\def\1{\bm{1}}

\DeclareMathAlphabet{\mathsfit}{\encodingdefault}{\sfdefault}{m}{sl}
\SetMathAlphabet{\mathsfit}{bold}{\encodingdefault}{\sfdefault}{bx}{n}

\usepackage{todonotes}
\usepackage{hyperref}
\usepackage{cleveref}
\crefname{appendix}{appendix}{appendices}
\Crefname{appendix}{Appendix}{Appendices}
\usepackage{url}
\usepackage{graphicx}
\usepackage{subcaption}
\usepackage{tikz}
\usepackage{array}
\usepackage{tabularray}
\usepackage{multirow}
\usepackage{wrapfig}
\usepackage{comment}
\usepackage{amsmath,amssymb,amsthm,mathtools} 
\usepackage{booktabs}
\usepackage{bm}
\usepackage{color}
\usepackage[dvipsnames]{xcolor}
\usepackage{fancyvrb}
\usepackage{algorithm,algpseudocode}
\usepackage{mathrsfs}
\DeclareMathSizes{7.5}{7}{5}{5}
\usepackage{microtype}
\usepackage{enumitem}
\usepackage{needspace}
\newcolumntype{V}{>{$\vcenter\bgroup\hbox\bgroup\strut}l<{\egroup\egroup$}}
\usetikzlibrary{arrows.meta,positioning,calc}
\setlist[itemize]{topsep=0pt,itemsep=0pt}
\algrenewcommand\algorithmicrequire{\textbf{In:}}
\algrenewcommand\algorithmicensure{\textbf{Out:}}
\hypersetup{
    colorlinks=true,
    linkcolor=black,
    citecolor=black,
    urlcolor=black,
}

\title{LATEC: Learned Axial TEnsor Compression for High-Dimensional Scientific Data}

\author{Hippolyte Verninas, Hadrien Hendrikx, Thomas Moreau \\
Université Paris-Saclay, Inria, CEA, Palaiseau, 91120, France \\
\texttt{\{hippolyte.verninas,hadrien.hendrikx,thomas.moreau\}@inria.fr}
}

\newcommand{\rebuttalupdate}[1]{#1}

\arxivfinalcopy
\begin{document}

\maketitle

\begin{abstract}
With the growing scale of scientific simulation and infrastructure, part of the burden has shifted from producing scientific data to analyzing and storing it. Compression is a key step to expanding storage capacity, but requires dedicated techniques to enforce user-specified pointwise error bounds that ensure the validity of downstream scientific workflows. To further push the limits of compression, novel learned codecs have been developed, in particular for images and videos, but these methods often optimize for aggregated distortion and fail to meet error bound standards. Moreover, these methods often target fixed field dimensionality, failing to account for many scientific data use cases, or at the expense of large adaptation or retraining costs.
In this work, we introduce the Learned Axial TEnsor Compression (LATEC) method\footnote{Code: \url{https://github.com/etyl/LATEC}}, an error-bounded codec with a dimension-agnostic learned predictor based on a GNN architecture. LATEC is pretrained on synthetic 2-D and 4-D fields, and evaluated with and without domain fine-tuning on 13 scientific fields in 3-D to 5-D. Results show that LATEC is consistently competitive with SZ3, achieving \rebuttalupdate{an 18.1\%} average improvement and greater consistency across all domains than other compressors. These results demonstrate that a shared learned predictor can serve as a single error-bounded codec across tensor dimensionalities, competitive with specialized compressors, while retaining pointwise error guarantees.
\end{abstract}

\section{Introduction}

Scientific simulation is central to understanding systems that are difficult, costly, or impossible to observe experimentally. As simulations scale, the limiting factor shifts from producing data to storing and analyzing it, because compute throughput grows faster than the I/O bandwidth and storage budgets available to write results out. A single density field of a gyrokinetic plasma simulation on a $64^5$ grid, three space and two velocity axes, already occupies 4\,GiB, and a production run relies on hundreds of such 5-D cells over many time steps and ensemble members. The shortfall is absorbed by discarding data: checkpoints are written at a coarser cadence, outputs are thinned, and ensembles are kept only in summary form, which limits post-hoc analysis and forces costly re-runs. Lossy compression is the standard way to keep more of this data, for archiving and transfer, for restart checkpoints, and to make repeated quantitative analyses practical~\citep{cappello2019usecases}.

What makes scientific compression different is not the size of the data but how it is reused. A compressed field may become the input to a restarted computation, a threshold query, an extremum search, or a derived physical diagnostic~\citep{cappello2019usecases,lee2022errorbounded}. None of these consumers inspect the field visually, and none tolerates an unknown worst case: a single corrupted value can move a detected extremum, flip a threshold test, or destabilize a restart, whatever the average error. Domain scientists therefore accept a compressed file as a substitute for the original only when the codec guarantees a pointwise bound $\lVert\mathbf{x}-\widehat{\mathbf{x}}\rVert_\infty\leq\varepsilon$ matching their own tolerance: it certifies every reconstructed value, and holds regardless of which analysis is run downstream. Aggregate criteria cannot play this role, since an RMSE constraint at the same tolerance still admits an isolated error as large as $\sqrt{n}\varepsilon$ in a field of $n$ values, which is why recent scientific benchmarks evaluate maximum-error compliance separately from average and structural distortion~\citep{reichelt2026climatebenchpress}. This bound is the minimal contract every such workflow shares; preserving specific derived quantities or topological features adds application-specific constraints on top of it~\citep{yan2023toposz}.

\begin{wraptable}{r}{0.53\textwidth}
\vspace{-0pt}
\centering
\scriptsize
\caption{Compressor comparison ($\dagger$: data-dependent).}
\label{tab:compressor-recap}
\vspace{-10pt}
\begin{tblr}{
  width = \linewidth,
  colspec = {X[1.1,l,m] *{4}{X[c,m]}},
  colsep = 1pt,
  row{1} = {font=\bfseries,halign=c},
  cell{1}{1} = {halign=l},
  hline{2} = {solid},
}
Compressor & Speed & \shortstack{Compression\\quality} & \shortstack{PWE\\bound} & \shortstack{Dim.\\agnostic} \\
SZ3 & $\checkmark\checkmark$ & $\checkmark$ & $\checkmark$ & \rebuttalupdate{$\times$} \\
SPERR & $\checkmark$ & $\checkmark\checkmark$ & $\checkmark$ & $\times$ \\
HPEZ & $\checkmark\checkmark$ & $\checkmark$ & $\checkmark$ & $\times$ \\
ZFP & $\checkmark\checkmark$ & $\times$ & $\checkmark$ & $\times$ \\
TTHRESH & $\checkmark$ & $\dagger$ & $\times$ & $\checkmark$ \\
\textbf{LATEC (ours)} & $\checkmark$ & $\checkmark\checkmark$ & $\checkmark$ & $\checkmark$ \\
\end{tblr}
\vspace{-14pt}
\end{wraptable}

Established error-bounded scientific compressors provide strong baselines without relying on application-specific training. \Cref{tab:compressor-recap} summarizes the main trade-offs among the evaluated families. The SZ family predicts values from previously reconstructed neighbors or multilevel interpolation, then quantizes residuals so that prediction quality affects the rate but not the pointwise guarantee~\citep{di2016fastsz,liang2022sz3}. Later variants improve GPU execution and select or compose predictors and parameters automatically~\citep{liu2024cusz,liu2022qoz,liu2023faz,liu2024hpez}. Other families make different structural choices: ZFP~\citep{lindstrom2014zfp} transforms independent blocks, MGARD~\citep{ainsworth2019multilevel, gong2023mgard} uses a multilevel decomposition, TTHRESH~\citep{ballester2019tthresh} exploits a higher-order tensor factorization, and SPERR~\citep{li2023sperr} combines wavelet coding with pointwise-error corrections. Their fixed analytic machinery makes these methods dependable and broadly applicable, but their coding efficiency depends on how well the chosen representation compacts each field.

Learned compression offers a way to exploit such structure, but existing architectures transfer poorly to scientific applications. Most progress has been driven by natural images and videos, where the reconstructed signal is ultimately judged through its rendering and codecs consequently optimize rate against an average or perceptual distortion. Their coding gains rest on nonlinear transforms paired with hierarchical or autoregressive entropy models~\citep{balle2017endtoend,balle2018variational,minnen2018joint}, refined since for practical use~\citep{tatwawadi2026pico} and, for video, through temporal prediction and learned motion representations~\citep{jia2025realtime}. Neither optimizes the criterion scientific workflows require, and both, like learned 3-D volume codecs, rely on convolutions whose kernels, feature layouts, and downsampling schedules assume a fixed number of dimensions. Learned scientific codecs similarly use blockwise or sparse autoencoders, constraint-aware reductions, 2-D super-resolution on slices, cross-field prediction, temporal graph models, or spatiotemporal latent models~\citep{liu2021aesZ,lee2022errorbounded,chung2025sparse,liu2023srnsz,jia2024neurlz,li2026extreme,li2025foundation}. Several add a residual correction stage to recover an error guarantee, but their learned component remains tied to a modality, dimensionality, or set of correlated variables. Many neural-field and autoencoder approaches are also fitted or trained anew for each dataset or domain~\citep{hayne2021neural,lu2021compressive,huang2023weather,dupont2022coinpp}. This cost is difficult to sustain when simulations expose many heterogeneous fields or when no representative training set is available.

A data-driven compressor for scientific tensors must therefore learn recurring structure without giving up the guarantees and broad applicability of predictive coding. Preserving native correlations requires exposing every tensor axis rather than flattening the data into slices, which privileges selected axes and discards correlations across the others. Yet naively extending a convolutional neighborhood of width $k$ to $D$ dimensions requires $k^D$ entries and a dimension-specific architecture. The compressor must also generalize without retraining for each new field. Sharing learned operations across axes, as in dimension-free axial networks~\citep{kim2025xnn}, provides a route to satisfying both requirements.

We introduce \emph{Learned Axial TEnsor Compression} (LATEC), an error-bounded predictive codec for scientific tensors of arbitrary dimensionality. LATEC uses learned prediction to improve coding efficiency while retaining the deterministic pointwise guarantee of residual-based scientific compressors. A single model, pretrained exclusively on synthetic 2-D and 4-D fields, transfers to 3-D, 4-D, and 5-D scientific data with no target-domain fine-tuning. Across 13 scientific fields in three to five dimensions, LATEC is the most rate-efficient compressor with a pointwise guarantee on \rebuttalupdate{12} fields. Relative to SZ3, it reduces the rate by \rebuttalupdate{18.1\%} on average, while its largest compression overhead on any field is \rebuttalupdate{13.1\%}.

\needspace{5\baselineskip}
Our contributions are:
\begin{itemize}[leftmargin=2em,beginpenalty=10000]
    \item \textbf{A dimension-agnostic GNN predictor}, which follows a coarse-to-fine schedule the decoder reproduces exactly, and applies without dimension-specific redesign, slicing, or flattening.
    \item \textbf{Scale-guided gating}, which uses the GNN-predicted residual scale to select between the learned and classical predictions, allowing fine-grained control while only adding a few bytes of overhead.
    \item \textbf{A synthetic field generator for domain-agnostic pretraining}, which samples local correlation, multiple scales, anisotropy, and sharp interfaces without emulating any governing equation. It is the sole pretraining signal, and also a stronger initialization than random when target-domain data is available for fine-tuning.
    \item \textbf{An open, extensible benchmark for error-bounded tensor compression}, which compares LATEC with five established compressors under both matched pointwise error and PSNR.
\end{itemize}

\section{Error-Bounded Learned Compression}
\label{sec:gnn-method}

\begin{figure}[t]
\centering
\includegraphics[width=.78\linewidth]{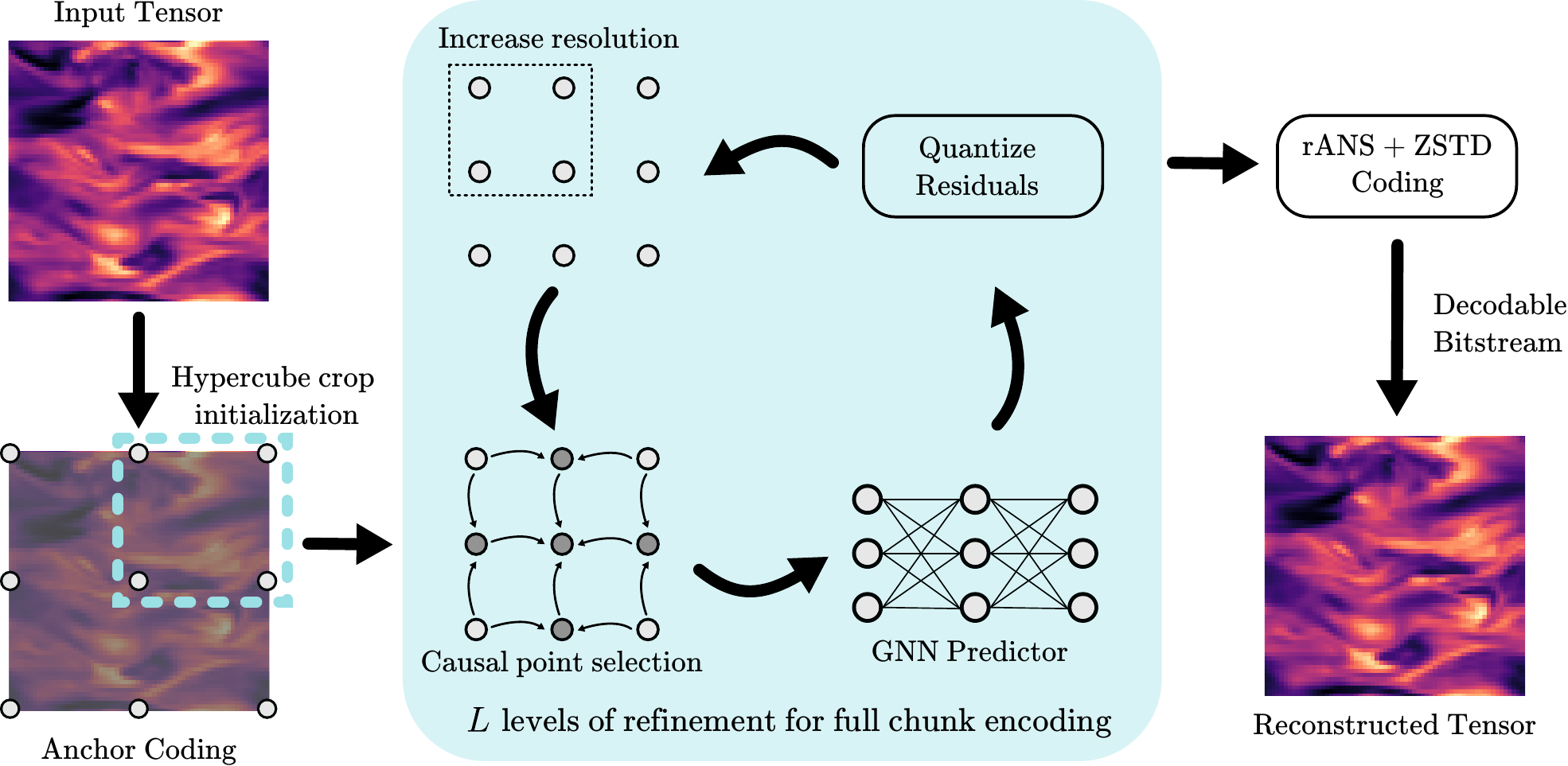}
\caption{The end-to-end codec. After the anchors are encoded, each chunk covers one anchor cell and undergoes $K$ levels of coarse-to-fine refinement. At every level, causally selected points are predicted by the gated GNN predictor, their residuals quantized under the prescribed error bound, and the  resolution is increased to predict intermediate values. The resulting symbols are coded with rANS and framed with Zstandard.}
\label{fig:overview}
\vspace{-6pt}
\end{figure}

LATEC separates the two roles of predictive compression: the coarse-to-fine residual coding inherited from the SZ family~\citep{di2016fastsz,liang2022sz3,liu2024hpez} provides the error guarantee (\Cref{sec:predictive-coding}), and learned prediction improves coding efficiency (\Cref{sec:axial-gnn})\rebuttalupdate{, with a per-point learned scale as in~\citet{balle2018variational}}. \Cref{fig:overview} gives an overview of the resulting pipeline. Throughout, $\mathbf{x}\in\mathbb{R}^{n_1\times\cdots\times n_D}$ is a scalar field on the lattice $\Omega=\prod_{j=1}^{D}\{1,\ldots,n_j\}$, $y_p\in[0,1]$ is the value at $p\in\Omega$ after normalization to the field's range, and the user-supplied range-relative pointwise-error bound becomes a tolerance $\varepsilon>0$ in these coordinates.

\subsection{Error-bounded predictive coding}
\label{sec:predictive-coding}

Predictive compression records, for each point of the field, not its value but how far that value departs from one that can already be computed from what has been decoded so far. The codec visits the points of $\Omega$ in a prescribed order: at point $p$ it forms a prediction $\hat{\mu}_p$ from previously reconstructed values, codes the residual $y_p-\hat{\mu}_p$, and reconstructs $\widehat y_p$, which then joins the context available to later points. Because predictions are computed from reconstructed values rather than original ones, the decoder reproduces the same sequence of states from the bitstream alone, which is what makes the loop closed. Compression comes from the residuals rather than the values: a predictor that fits the field concentrates them near zero, and the entropy coder then spends few bits per point. The tolerance $\varepsilon$ enters only where the residuals are quantized, which is what makes the scheme error-bounded.

\textbf{Anchors and the coarse-to-fine schedule.}
The visiting order must give every prediction a context the decoder already holds. LATEC adopts the dyadic, coarse-to-fine order of SZ-family multidimensional interpolation. The \emph{anchor} lattice, the sublattice of stride $s_K=2^K$ along every axis, is coded first and with no prediction at all: its values are quantized and coded directly. Anchors are the only values coded without context, and they bootstrap all the others. The $K$ levels that follow, indexed by their stride $s_i=2^i$ down to $s_0=1$, each halve the stride and predict the midpoints between values the previous level made available. Within a level, points whose neighbors lie along a single axis are coded before those requiring several axes, which splits each level into $2D-1$ \emph{sub-stages} (\Cref{app:causal-schedule}), giving a total of $(2D-1)K$ \emph{stages}. The anchor lattice also partitions the domain into \emph{chunks}, cells of edge $s_K$ that these levels are coded in one at a time (\Cref{sec:chunking}); one stage restricted to one chunk is a \emph{chunk-stage}, the unit at which the gate of \Cref{sec:axial-gnn} is chosen. As errors at coarse levels propagate into every later prediction, these levels use a tighter tolerance than requested: level $i$ codes at $\varepsilon_i=\varepsilon\rho^{i}$, with $0<\rho\leq1$, so the anchors code at the tightest tolerance $\varepsilon_K$ and the finest level at $\varepsilon_0=\varepsilon$.

\textbf{Quantization and the error guarantee.}
The prediction supplies a reconstruction center while residual quantization enforces the bound. The encoder converts the residual to the nearest quantization index,
\begin{equation}
 q_p=\operatorname{round}\left(\frac{y_p-\hat{\mu}_p}{2\varepsilon_i}\right),
 \label{eq:quant-index}
\end{equation}
and, for an in-range index, both sides reconstruct the value with exactly the arithmetic used for encoder-side verification:
\begin{equation}
 \widehat y_p= \hat{\mu}_p + 2 \varepsilon_i q_p.
 \label{eq:reconstruction}
\end{equation}
Rounding to the nearest level of a grid of spacing $2\varepsilon_i$ leaves an error of at most $\varepsilon_i$, and the encoder verifies $|y_p-\widehat y_p|\leq\varepsilon_i$ explicitly for every quantized value. Note that \textbf{this guarantee is a property of the quantizer},  independent of $\hat{\mu}_p$: the quantization grid is centered on the prediction, so a poor estimate yields a large index and costs bits, but cannot increase the error. The only values left uncovered are those whose index exceeds the coded range or whose verification fails, coded separately (\Cref{app:error-control}).

\textbf{Entropy coding.}
The retained indices are coded with range asymmetric numeral systems (rANS)~\citep{duda2013ans} and framed with Zstandard~\citep{collet2021zstd}. Alongside $\hat{\mu}_p$, the predictor returns a scale $\hat\beta_p$ parameterizing a Laplace distribution over the residual at $p$, which serves as the entropy model. \Cref{app:entropy} details it, and \Cref{fig:ablation-entropy} compares it with Huffman coding and a fixed scale.

\subsection{Axial GNN predictor}
\label{sec:axial-gnn}

\begin{figure*}[t]
\centering
\includegraphics[width=\linewidth]{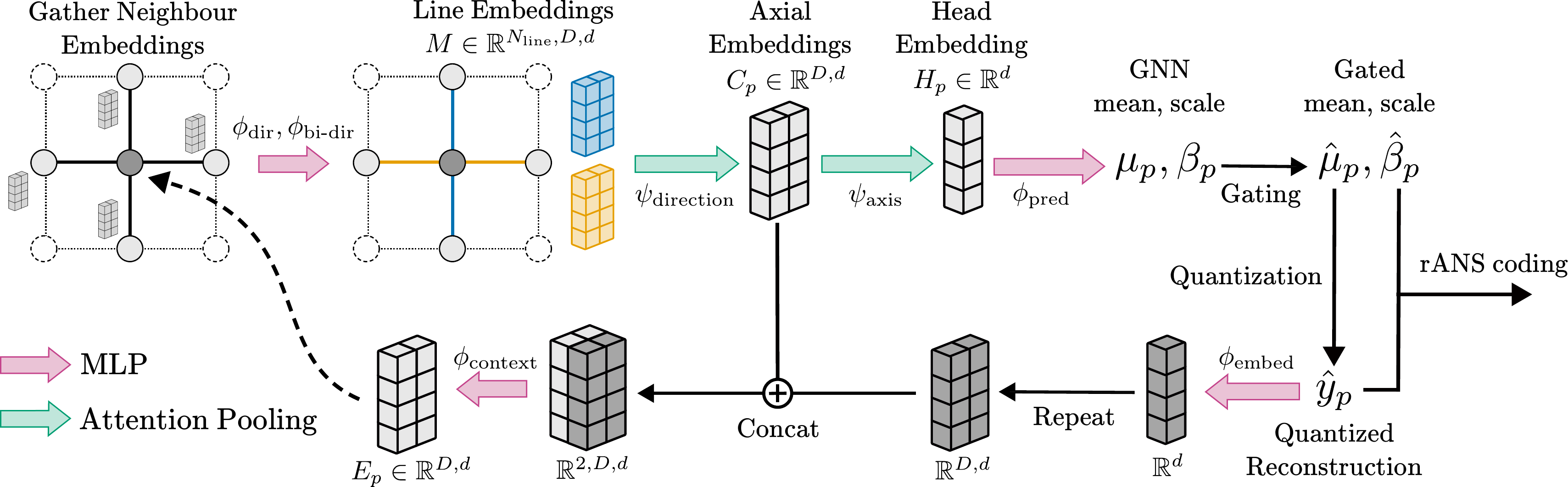}
\caption{Causal axial GNN at one prediction stage. Line embeddings are pooled into one embedding per axis (two are shown for the 2-D case), then into the head that predicts $(\mu_p,\beta_p)$, which gives $(\hat{\mu}_p,\hat{\beta}_p)$ after gating. The residual is quantized and entropy-coded, and the reconstructed value yields the embedding that updates the state $E$.}
\label{fig:architecture}
\end{figure*}

Classical codecs use a fixed interpolation scheme, with coefficients chosen in advance and a shape tied to the number of axes. LATEC replaces the predictor with a GNN which builds an embedding for each reconstructed point and reuses these embeddings to predict the mean and scale for the points coded next. \rebuttalupdate{Note that our GNN is not a deep GNN. It gathers information from each point's neighborhood, and process each axis with a shared small 1D MLP. These embedding along each dimension are then aggregated using attention pooling.} Because the schedule is unchanged, \textbf{encoder and decoder still see identical causal context} in any number of dimensions. \Cref{fig:architecture} shows one prediction step: line embeddings formed from the causal neighbors, pooling per axis and then across axes, a head that outputs $(\mu_p,\beta_p)$, and a state update from the reconstructed value.

When coding a point $p \in \Omega$, the GNN may use only values reconstructed at earlier stages. It reaches them through a state $E\in\mathbb{R}^{|\Omega|\times D\times d}$, holding one embedding of width $d$ per already-coded point and per lattice axis. The causal context of $p$ is the nearest such point on either side of each of the $D$ axis-aligned lattice lines through $p$. For each available line $\ell$ and axis $j$, the GNN combines the neighbor embeddings $E^\pm_{\ell,j}$ with context embeddings $Z^\pm_{\ell,j}$ that encode their direction and geometry. A line provides two-sided context when a reconstructed point is available on each side of $p$, and one-sided context when only a single point is available, as at a domain edge. Separate MLPs handle these two cases:
\begin{equation}
 M_{\ell,j}=\begin{cases}
 \phi_{\rm bi\text{-}dir}([E^-_{\ell,j},Z^-_{\ell,j},E^+_{\ell,j},Z^+_{\ell,j}]),
 &\text{two-sided},\\
 \phi_{\rm dir}([E^\sigma_{\ell,j},Z^\sigma_{\ell,j}]),
 &\text{one-sided}.
 \end{cases}
 \label{eq:line-embedding}
\end{equation}

Each available line $\ell$ produces one embedding $M_{\ell,j}$ for every axis $j$. Let $\mathcal{I}_p$ index the available lines through $p$, with $|\mathcal{I}_p|\leq D$. Following graph-attention pooling~\citep{velickovic2018gat}, learned single-query maps $\psi_{\rm direction}$ and $\psi_{\rm axis}$ first aggregate line embeddings separately for each axis, then aggregate the resulting per-axis contexts:
\begin{equation}
 C_{p,j}=\psi_{\rm direction}(\{M_{\ell,j}\}_{\ell \in \mathcal{I}_p}),\qquad
 H_p=\psi_{\rm axis}(\{C_{p,j}\}_{j=1}^D).
 \label{eq:pools}
\end{equation}
Pooling up to $D$ line embeddings for each of $D$ axis slots, then the $D$ resulting axis contexts, costs $\mathcal{O}(D^2)$ per query at fixed embedding width. Because the line-embedding and pooling maps are shared across axes, the learned parameter count is independent of $D$, consistent with the dimension-free principle of axial neural networks~\citep{kim2025xnn}.

The prediction head also receives the stage tolerance $\varepsilon_i$. Conditioned on $H_p$ and $\log_2\varepsilon_i$, \textbf{the head returns a mean and Laplace scale},
\begin{equation}
 (\mu_p,\log_2\beta_p)=\phi_{\rm pred}([H_p,\log_2\varepsilon_i]),\quad
 \mu_p\in[0,1],\quad \beta_p\in[\varepsilon_i/16,4096\varepsilon_i].
 \label{eq:head}
\end{equation}
Both the conditioning and the bounds are relative to $\varepsilon_i$, and the scale is predicted as $\log_2\beta_p$: what matters for coding is the residual measured in quantization steps, not its absolute size. A given ratio $\beta_p/\varepsilon_i$ then means the same at every level and tolerance, so one trained head serves the whole schedule.

After quantization reconstructs $\widehat y_p$, \textbf{encoder and decoder finalize the state at $p$ by mixing that reconstructed value into its per-axis context}, $E_{p,j}\leftarrow\phi_\text{context}(C_{p,j},\phi_\text{embed}(\widehat y_p))$ for every axis $j$, with the learned maps detailed in \Cref{app:chunk-state}. Using $\widehat y_p$ rather than $y_p$ makes this closed-loop state reproducible at the decoder. Reusing these finalized embeddings at later stages propagates context through the hierarchy, even though each prediction uses a single graph-aggregation step.

\subsection{Scale-guided gating}
\label{sec:gating}

The GNN predicts better than interpolation on average, but not everywhere: on locally smooth regions of some fields, a classical interpolant can be more accurate. LATEC therefore keeps three classical candidates $\{\mu^{(a)}_p\}_{a=1}^3$ alongside the GNN mean: cubic interpolation, and linear interpolation along the first or last axis on which $p$ is a midpoint. Like the GNN, all three use only reconstructed values, so the decoder can evaluate them.

Which predictor wins changes across fields, stages, and regions, so a fixed choice is suboptimal, while a per-point selection flag would be too costly.
One candidate $a$ is fixed for a whole chunk-stage; what varies from point to point is only whether it is used in place of the GNN mean, \textbf{depending on the gating decision $g_p$ made from the predicted scale $\beta_p$}.
The decision function is parametrized for a given chunk-stage by a threshold $\varepsilon_i2^T$, with $T\in\{2,\ldots,13\}$, on the same $\varepsilon_i$-relative grid as $\beta_p$, and a sign $d \in \{+1,-1\}$, which determines the direction of the threshold. The gate reads:
\begin{equation}
    g_p=
    \mathbf{1}[d \beta_p<d\varepsilon_i2^T]
\qquad
 \widehat\mu_p=g_p\,\mu^{(a)}_p+(1-g_p)\,\mu_p,
 \qquad
 \widehat\beta_p=2^{-s g_p}\,\beta_p,
 \label{eq:gate}
\end{equation}
At the replaced points, the scale $\beta_p$ was predicted for the GNN residual, so it would overestimate the residual given by the classical interpolant. The gate therefore multiplies it by a factor of $2^{-s}$, $s\in\{0,2,4,6\}$, so that the entropy model matches the smaller residual of the classical mean.

A gate is therefore described by four discrete choices, $(a,d,T,s)$. For each chunk-stage, the encoder estimates the code length of the stage's residuals under the GNN alone and under each of the 288 combinations, and keeps the cheapest. The gate stays off unless some combination is strictly cheaper than the GNN alone (\Cref{app:gate}). The selected parameters $(a,d,T,s)$ are transmitted as one 16-bit word per chunk-stage. The decoder, which recomputes $\beta_p$, then rebuilds the mask $g_p$ on its own: the choice of predictor is made \textbf{per point}, whereas only one word is transmitted \textbf{per chunk-stage}. \Cref{fig:ablation-predictor,fig:gate-selection} compare this predictor with fixed interpolation and show the gate's selections.

\rebuttalupdate{\textbf{Per-chunk-stage scale shift.} Once a chunk-stage's symbols are fixed, the entropy coder can still correct a scale that is systematically off on that stage, for instance on fields far from the pretraining distribution, by coding each point under
\begin{equation}
 \widetilde\beta_p=\eta^{\kappa}\,\lfloor\widehat\beta_p\rceil,
 \label{eq:scale-shift}
\end{equation}
where $\lfloor\cdot\rceil$ rounds to the coder's scale grid of step $\eta$ (\Cref{app:entropy}) and $\kappa\in[-21,21]$ is one integer per chunk-stage, chosen to minimize its bitrate and transmitted in its stage descriptor. Unlike the gate shift $s$, $\kappa$ applies to every point of the chunk-stage.}

\subsection{Chunked execution for bounded memory}
\label{sec:chunking}

\begin{wrapfigure}{r}{0.28\textwidth}
\vspace{-14pt}
\centering
\includegraphics[width=\linewidth]{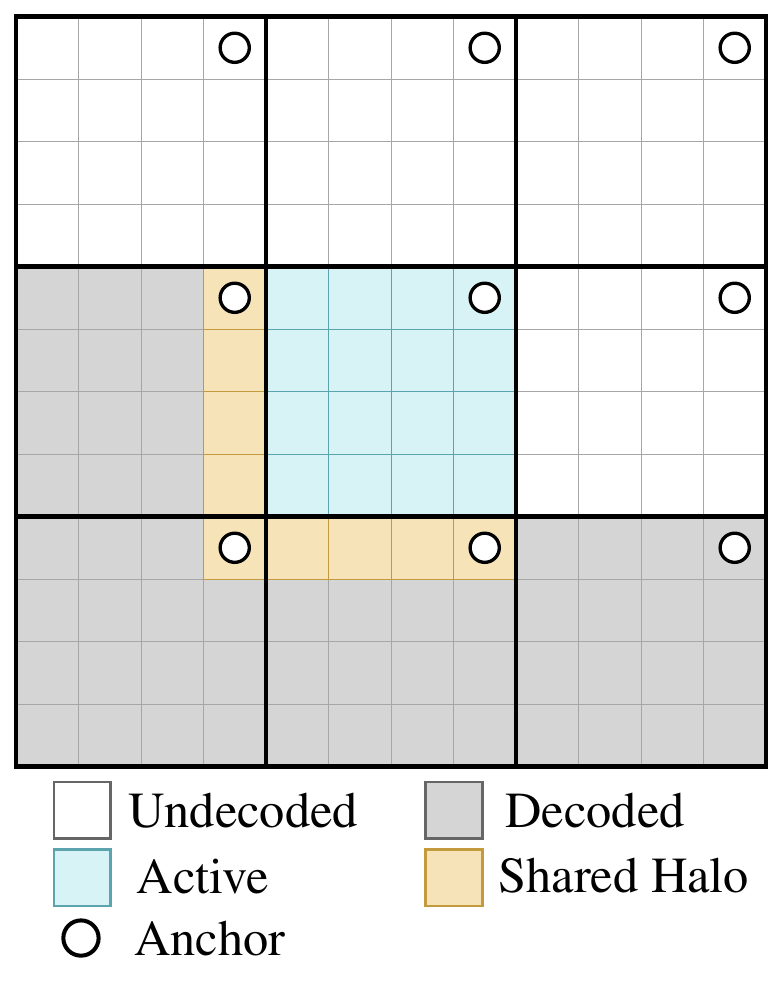}
\vspace{-20pt}
\caption{Chunking in 2-D.}
\label{fig:chunking}
\vspace{-18pt}
\end{wrapfigure}

Learned prediction pays for its coding gain with per-point state that a classical codec does not carry. At the sizes where compression matters most, this state decides whether the codec can run at all. The state update finalizes one embedding of width $d$ for every point and every axis, so retaining that field for a whole tensor would cost $\mathcal{O}(|\Omega|Dd)$: at the deployed width $d=64$ in float32, $64D$ times the size of the input, or 1.25\,TiB for the 4\,GiB $64^5$ field of the introduction. Coding one chunk at a time and discarding its embeddings on completion instead gives active storage $\mathcal{O}(VDd+|\Omega|)$, a neural term over the extended chunk volume $V$ plus the scalar reconstruction buffer any codec must hold. The neural term $VDd$ is then fixed by the configured chunk size and independent of the input, so \textbf{total memory grows with one scalar per point} rather than with $Dd$ of them. \rebuttalupdate{\Cref{sec:experiments} measures 2.0-2.9\,MiB per input MiB from 2-D to 6-D.}

Chunks are visited in raster order, and each inherits its low-coordinate faces from already reconstructed neighbors, as shown in \Cref{fig:chunking}. This partition gives every point a single coding owner while supplying the boundary values later chunks need. Halo embeddings are rebuilt from within the current chunk rather than carried across the seam, so no learned state is retained between chunks (\Cref{app:chunk-state}).

\subsection{Domain-agnostic training on synthetic fields}
\label{sec:training-stream}

\textbf{Objective.}
The predictor is trained to minimize the bitrate it determines: the code length, in bits, that the entropy model parameterized by its predicted mean and scale assigns to each quantization index of \Cref{eq:quant-index}. Two substitutions make that cost cheap to evaluate and differentiable. First, the closed loop of deployment is replaced by teacher forcing: training follows the causal stage order but supplies later stages with $\operatorname{clip}(y_p+\eta_p,0,1)$, where $\eta_p\sim\mathcal{U}[-\varepsilon,\varepsilon]$ has the same support as the reconstruction error it stands in for, so the predictor sees imperfect context without the encoder running. Second, the mass of the cell selected by $\operatorname{round}$ is replaced by the mass of the width-$2\varepsilon$ interval centered on the continuous residual, which takes the rounding out of the graph and leaves gradients on both the mean and the scale. Writing $F_{\rm Lap}(\cdot\,;0,\beta_p)$ for the zero-mean Laplace CDF with predicted scale $\beta_p$, the objective over the set $\mathcal{H}$ of non-anchor positions is
\begin{equation}
 \mathcal{L}=\frac{1}{|\mathcal{H}|}\sum_{p\in\mathcal{H}}
 -\log_2\!\left[
 F_{\rm Lap}(y_p-\mu_p+\varepsilon;0,\beta_p)-F_{\rm Lap}(y_p-\mu_p-\varepsilon;0,\beta_p)
 \right].
 \label{eq:training-objective}
\end{equation}
For $|y_p-\mu_p|\geq\varepsilon$ a logarithmic continuation of the Laplace tail limits extreme-residual gradients. The gate and the deployed entropy-table selection are decisions taken by the encoder at inference and are excluded from training. \Cref{app:training-details} gives the training details.

\textbf{Synthetic generation.}
We train the predictor exclusively on fields synthesized on the training device, as done for tabular and time-series foundation models~\citep{hollmann2025tabpfn,qu2025tabicl,xie2026cauker}, which removes the need for a representative target-domain corpus, permits unlimited online resampling, and rules out target-domain leakage. The generator emulates no governing equation, and transfer is not expected to follow from any resemblance between its output and a simulated field. What the predictor sees at each query is a causal neighborhood a few points wide, and what sets the residual there is local structure: how fast the field decorrelates, whether it decorrelates at the same rate along every axis, how its values are distributed, and whether an interface crosses the neighborhood. The generator is built to span those properties rather than to imitate any physics.

Each realization draws three independent choices, one per property, listed in \Cref{tab:synthetic-realizations}. The spectrum is either smooth, white noise filtered by an anisotropic Gaussian spectrum at two correlation scales, or turbulent, a radial power law repeatedly warped by a smooth random displacement; between them they supply local correlation at several scales and anisotropy. A monotone value warp then varies the marginal distribution, and a number of random hyperplane jumps inserts sharp interfaces. The choices give $2\times3\times4=24$ discrete construction types with continuous parameters resampled within each.

A batch mixes $128\times128$ 2-D and $16^4$ 4-D fields, resampled at every optimization step, in numbers that give the two dimensionalities equal weight in the loss. Half the 2-D steps split the field into two chunks with a randomized split axis and traversal order, so the predictor is also trained under the halo conditions of \Cref{sec:chunking} rather than on whole fields alone. \Cref{app:training-details} gives the equations, parameter distributions, boundary conditions, and normalization.

\begin{figure*}[t]
\centering
\includegraphics[width=.9\linewidth]{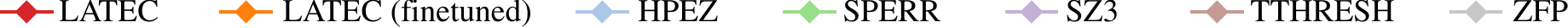}\\[1.5mm]
\includegraphics[width=\linewidth]{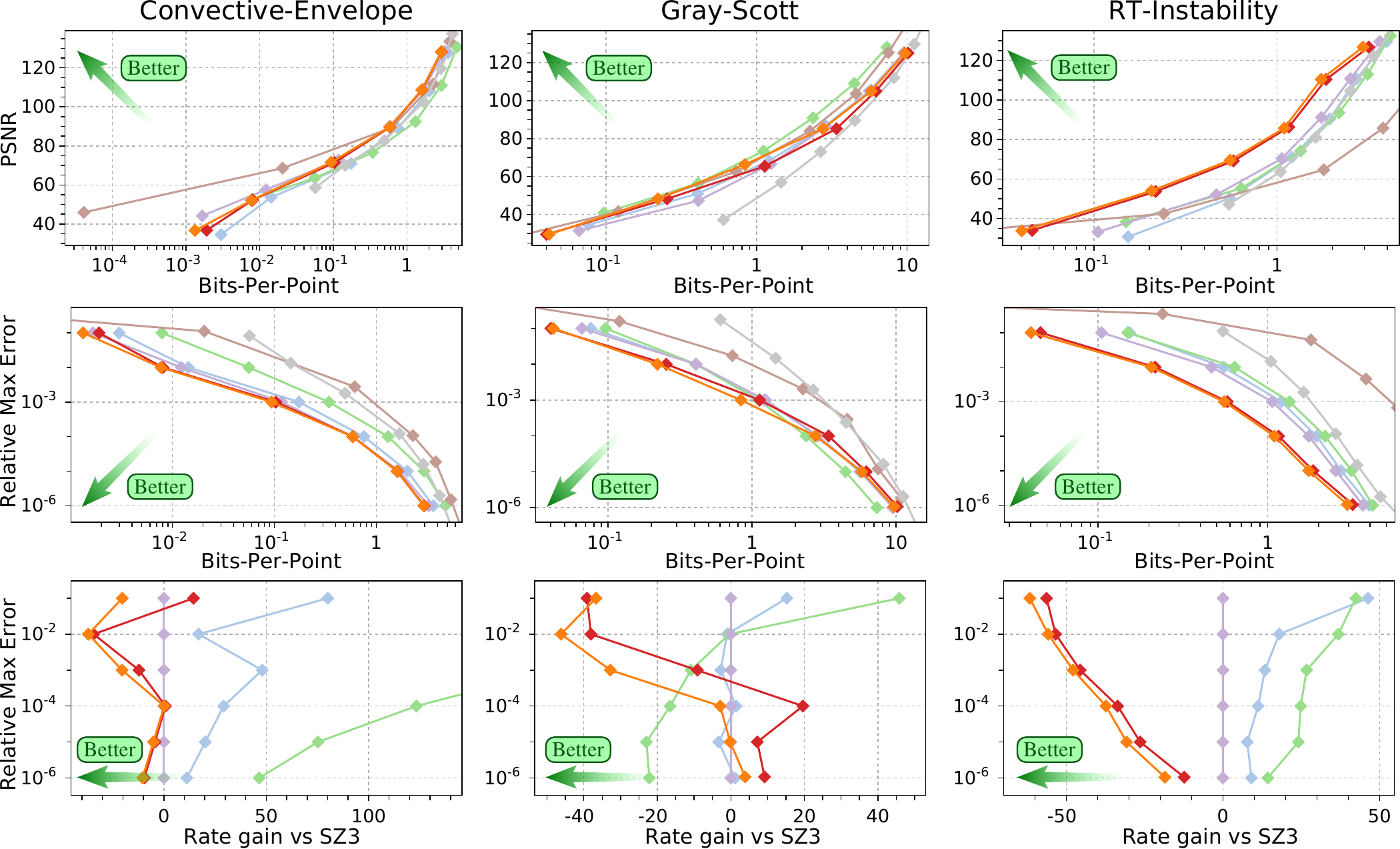}
\caption{Rate-quality results on three datasets from The Well. We compare PSNR and RME against BPP and the relative BPP difference with SZ3.}
\label{fig:rate-distortion}
\vspace{-2pt}
\end{figure*}

\section{Cross-dimensional evaluation}
\label{sec:experiments}

\textbf{An open benchmark for error-bounded tensor compression.}
We release the evaluation as a Benchopt benchmark~\citep{moreau2022benchopt}, where datasets, compressors, and metrics are independent components, so a new field or codec plugs in without touching the rest.\footnote{\label{fn:benchmark}\url{https://github.com/benchopt/benchmark_scidata_compression}} It covers 13 scalar fields in three to five dimensions, drawn from SDRBench~\citep{zhao2020sdrbench}, The Well~\citep{ohana2024well}, the 5-D gyrokinetic simulations of GyroSwin~\citep{paischer2026gyroswin} and Gysela~\citep{grandgirard2006gysela}, and 5-D synthetic data (\Cref{app:synthetic-5d}), listed with their shapes and sizes in \Cref{tab:datasets}. The 5-D fields test whether a codec retains correlations across more than three axes without a privileged slicing axis. The predictor is pretrained on 2-D and 4-D synthetic fields only (\Cref{sec:training-stream}), so 3-D and 5-D are unseen.

For each field and operating point we report compressed size in bits per point (BPP), the achieved range-relative maximum error (RME), and the peak signal-to-noise ratio (PSNR), defined in \Cref{app:metrics}. We compare LATEC against SZ3, SPERR, HPEZ, ZFP, and TTHRESH, each under its own error-control configuration. Because TTHRESH controls aggregate error rather than RME, its matched-RME values are estimates, included in the ranking but marked with an asterisk. We also evaluated the learned CAESAR compressor~\citep{li2025foundation}, but it performed poorly under our evaluation regime, so we report its results separately in \Cref{app:caesar}. Other learned methods, including AE-SZ~\citep{liu2021aesZ} and SRN-SZ~\citep{liu2023srnsz}, are excluded because their code is not publicly available.
We evaluate the pretrained model on all 13 fields and, wherever a separate training split exists, namely each evaluated The Well dataset\rebuttalupdate{ and GyroSwin}, a checkpoint fine-tuned on that split alone and evaluated on a held-out test trajectory\rebuttalupdate{, and for the single-run Gysela, on disjoint crops} (\Cref{sec:training-stream} and \Cref{app:training-details}).

\begin{table*}[t]
\centering
\fontsize{9}{9}\selectfont
\caption{Average percentage BPP difference from SZ3 across matched relative maximum error (RME) targets for \colorbox{blue!8}{3-D}, \colorbox{green!10}{4-D}, and \colorbox{orange!12}{5-D} datasets. Negative values are rate savings and lower is better. * marks estimated RME values because the compressor provides no bound, and ``/'' marks a dataset \rebuttalupdate{without fine-tuning data, hence without a fine-tuned checkpoint}. \textbf{Bold} is best, \underline{underlined} second-best.}
\label{tab:error-bpp-saving}
\vspace{-4pt}
\newcommand{\rotateheader}[1]{\makebox[0pt][c]{\rotatebox[origin=lb]{90}{\strut #1}}}
\begin{tblr}{
  colspec = {l*{13}{r}},
  colsep = 1.5pt,
  row{1} = {halign=c,valign=b,belowsep=2pt,cmd=\rotateheader},
  cell{1}{1} = {cmd={},halign=l},
  column{2-6} = {bg=blue!8},
  column{7-11} = {bg=green!10},
  column{12-14} = {bg=orange!12},
  hline{2} = {solid},
}
Method & \shortstack[l]{Euler-Multi-\\Quadrants} & Gray-Scott & \shortstack[l]{Hurricane\\ISABEL} & Miranda & S3D & \shortstack[l]{Convective-\\Envelope} & \shortstack[l]{Gravity-\\Cooling} & MHD-64 & \shortstack[l]{RT-\\Instability} & \shortstack[l]{Supernova-\\Explosion} & GyroSwin & Gysela & Synthetic 5D \\
SZ3 & \underline{0} & 0 & \underline{0} & 0 & 0 & 0 & 0 & 0 & 0 & 0 & 0 & 0 & \underline{0} \\
ZFP & 976.8 & 525.6 & 425.2 & 2148.0 & 4580.1 & 2366.1 & 4827.5 & 493.3 & 260.5 & 4369.6 & \rebuttalupdate{928.4} & 2386.1 & 209.8 \\
SPERR & 26.0 & -5.0 & 108.1 & 4.7 & 0.3 & 195.1 & 20.7 & 5.7 & 28.7 & 138.7 & \rebuttalupdate{13.6} & 20.9 & 14.0 \\
HPEZ & \textbf{-4.4} & 2.9 & 24.6 & \underline{-4.8} & 2.8 & 34.6 & 19.8 & 10.3 & 17.6 & 84.5 & \rebuttalupdate{27.0} & 177.6 & 12.8 \\
TTHRESH* & 890.8 & 112.5 & 1293.2 & 89.6 & \underline{-14.8} & 654.3 & 265.5 & 120.5 & 547.9 & 2912.9 & \rebuttalupdate{38.1} & \textbf{-31.4} & 69.5 \\
\textbf{LATEC} & \rebuttalupdate{13.1} & \rebuttalupdate{\underline{-8.6}} & \rebuttalupdate{\textbf{-16.5}} & \rebuttalupdate{\textbf{-15.8}} & \rebuttalupdate{\textbf{-25.7}} & \rebuttalupdate{\underline{-7.5}} & \rebuttalupdate{\underline{-18.5}} & \rebuttalupdate{\underline{-21.2}} & \rebuttalupdate{\underline{-38.1}} & \rebuttalupdate{\underline{-17.9}} & \rebuttalupdate{\underline{-27.4}} & \rebuttalupdate{-3.0} & \rebuttalupdate{\textbf{-48.6}} \\
\shortstack[l]{\textbf{LATEC}\\(fine-tuned)} & \rebuttalupdate{6.0} & \rebuttalupdate{\textbf{-19.4}} & / & / & / & \rebuttalupdate{\textbf{-15.5}} & \rebuttalupdate{\textbf{-32.2}} & \rebuttalupdate{\textbf{-23.4}} & \rebuttalupdate{\textbf{-42.1}} & \rebuttalupdate{\textbf{-26.8}} & \rebuttalupdate{\textbf{-30.8}} & \rebuttalupdate{\underline{-16.8}} & / \\
\end{tblr}
\vspace{-8pt}
\end{table*}

\begin{wrapfigure}[15]{r}{0.29\textwidth}
\vspace{-10pt}
\centering
\includegraphics[width=\linewidth]{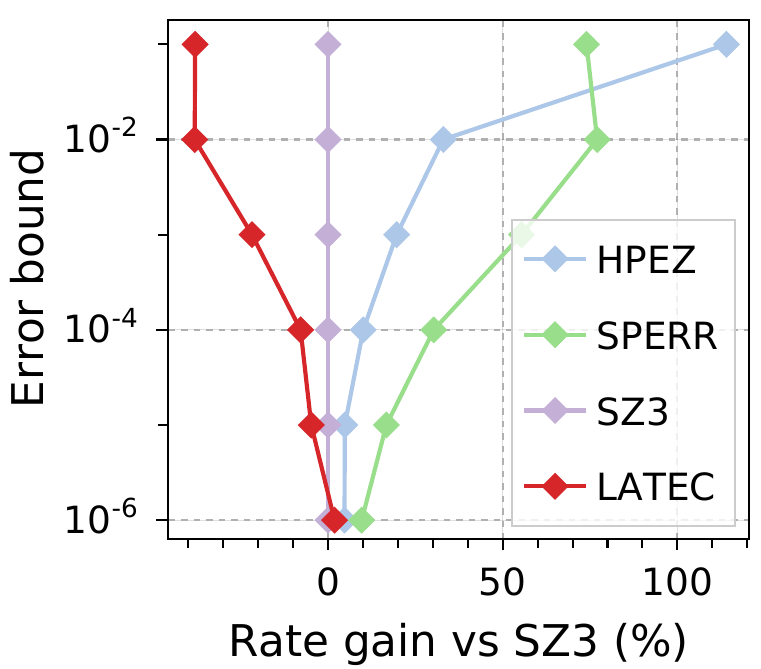}
\vspace{-18pt}
\caption{Dataset-averaged rate difference from SZ3 by RME target.}
\label{fig:global-savings}
\vspace{-10pt}
\end{wrapfigure}

\paragraph{Rate-error results.}
At matched RME, pretrained LATEC is the most rate-efficient compressor on \rebuttalupdate{11} of the 13 fields, using fewer bits than each of SZ3, ZFP, SPERR, HPEZ and TTHRESH (\Cref{fig:rate-distortion,tab:error-bpp-saving}). Among compressors with a pointwise guarantee, which excludes the estimated TTHRESH entries, it is the most rate-efficient on \rebuttalupdate{12} fields. Pairwise, it beats ZFP and \rebuttalupdate{SPERR on all 13 fields, and HPEZ, SZ3, and the estimated TTHRESH entries on 12, losing only on Euler-Multi-Quadrants (to SZ3 and HPEZ) and on Gysela (to TTHRESH).} With the fine-tuned checkpoint on the \rebuttalupdate{nine} adapted fields and the pretrained checkpoint elsewhere, LATEC is best on \rebuttalupdate{11} of the 13 fields. \rebuttalupdate{On the real 5-D fields, it saves $27.4\%$ on GyroSwin and $3.0\%$ on Gysela relative to SZ3 ($30.8\%$ and $16.8\%$ fine-tuned)} while coding them in their native coordinates; \rebuttalupdate{only TTHRESH also receives the native shape there, whereas SZ3, ZFP, HPEZ, and SPERR must group axes} (\Cref{app:synthetic-5d}).

\paragraph{Consistency across fields.}
LATEC's aggregate advantage is not confined to a single error regime. It achieves the best dataset-averaged rate from RME $10^{-1}$ through $10^{-5}$. However its lead narrows as the bound tightens and SZ3 has a small advantage at $10^{-6}$ (\Cref{fig:global-savings}). HPEZ and SPERR, in contrast, incur an average overhead at every target, and their gap from SZ3 widens markedly as the bound loosens. Averaged over the fields of \Cref{tab:error-bpp-saving}, LATEC is the only evaluated compressor whose mean rate falls below SZ3's, with a \rebuttalupdate{18.1\%} mean and \rebuttalupdate{17.9\%} median saving, and its worst field costs \rebuttalupdate{13.1\%} more than SZ3. The other compressors have much larger worst cases, from HPEZ's 177.6\% on Gysela to 2912.9\% for the estimated TTHRESH entries on supernova explosion, and ZFP averages \rebuttalupdate{1884.4\%}.

\begin{figure}[t]
\centering
\includegraphics[width=.86\linewidth]{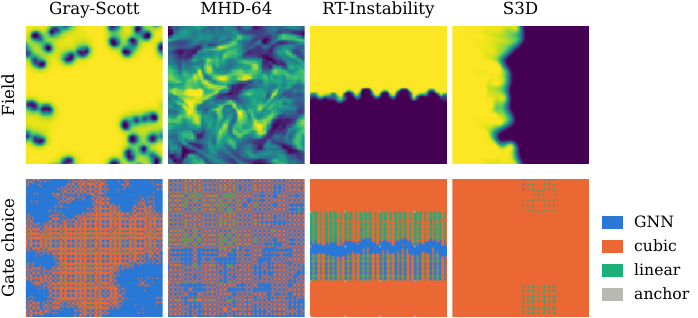}
\vspace{-4pt}
\caption{\rebuttalupdate{Predictor selected by the gate on one 2-D slice of four fields at RME $10^{-4}$.}}
\label{fig:gate-choice}
\vspace{-8pt}
\end{figure}

\begin{wrapfigure}[13]{r}{0.4\textwidth}
\vspace{-16pt}
\centering
\includegraphics[width=\linewidth]{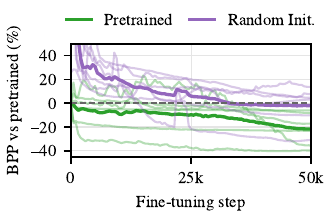}
\vspace{-18pt}
\caption{Held-out BPP during fine-tuning on The Well, relative to the frozen pretrained model.}
\label{fig:finetuning}
\vspace{-10pt}
\end{wrapfigure}

\paragraph{Rate-distortion results.}
The PSNR curves in \Cref{fig:rate-distortion} provide a secondary view of average reconstruction error at a given rate. At matched PSNR, pretrained LATEC uses fewer bits than SZ3 on \rebuttalupdate{9} of the 13 fields and, with the fine-tuned checkpoint, is the best method on \rebuttalupdate{6} of them; SPERR, HPEZ, and especially TTHRESH remain stronger on several datasets (\Cref{tab:distortion-bpp-saving} in \Cref{app:psnr-table}). The two views differ for a reason shared by the whole prediction-and-quantization family: uniform quantization at spacing $2\varepsilon_i$ spreads the error budget evenly over every sample, which makes the $L^\infty$ contract inexpensive to satisfy but is not how bits would be allocated to minimize squared error, so transform methods truncating against an $L^2$ criterion are favored by PSNR. The matched-RME comparison in \Cref{tab:error-bpp-saving} remains the primary evaluation of compression rate under the requested pointwise bound.

\begin{wrapfigure}[17]{l}{0.42\textwidth}
\vspace{-20pt}
\centering
\includegraphics[width=\linewidth]{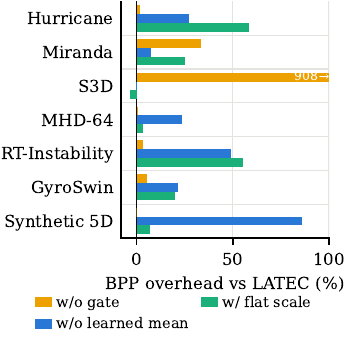}
\vspace{-20pt}
\caption{\rebuttalupdate{Ablation of LATEC components at RME $10^{-3}$, one per bar. The off-scale S3D bar is clipped.}}
\label{fig:ablation-main}
\vspace{-10pt}
\end{wrapfigure}

\paragraph{Effect of the synthetic pretraining.}
Fine-tuning lowers LATEC's BPP on all \rebuttalupdate{nine} adapted datasets, at matched RME and at matched PSNR (\Cref{tab:error-bpp-saving,tab:distortion-bpp-saving}). These gains come from the pretrained initialization rather than from target data alone: training the same architecture from random initialization on each of the seven evaluated datasets from The Well gives a substantially higher average bitrate early in training, and the gap persists at the final checkpoint (\Cref{fig:finetuning}).

\rebuttalupdate{\paragraph{Ablations.}
\Cref{fig:ablation-main} removes one learned component of LATEC at a time (setup in \Cref{app:ablations}). Compared with a classical gate that selects between cubic and linear interpolation, with the same chunk-stage signaling and the same learned scale, the GNN mean saves 7-33\% of the bits on the real fields other than S3D, and 46\% on Synthetic 5D; on the smooth S3D field, the classical gate is 0.3\% cheaper. Keeping the mean and the gate but coding the same quantized residuals under the best flat scale of each chunk-stage, instead of the pointwise learned scale, costs 20-58\% more bits on Hurricane, Miranda, RT-Instability and GyroSwin, and 4-7\% on MHD-64 and Synthetic 5D. Only on S3D, where the GNN is almost never selected, is the flat scale cheaper, by 3\%. Conversely, the GNN alone fails on smooth 3-D data: on S3D it uses 8.7 times as many bits as SZ3. The gate removes 90\% of these bits on S3D and 25\% on Miranda, while on the other fields, where the GNN alone already beats SZ3, it saves a further 0-5.3\%. The selection is spatially structured (\Cref{fig:gate-choice}): the classical interpolants are chosen in smooth regions, where they are already accurate, and the GNN in nonlinear or turbulent ones. On Gray-Scott the GNN takes the spots, where the nonlinear reaction creates sharp fronts, and cubic interpolation the smooth background; on RT-Instability the GNN follows the turbulent mixing interface, linear interpolation the rows next to it, and cubic interpolation the pure-fluid regions. Each chunk-stage transmits a single 16-bit rule, so this pointwise selection follows from the variation of $\beta_p$ alone, without per-point flags. Scaling with $D$ up to 6-D, per-input training, and model width are studied in \Cref{app:dimensionality,app:per-input,app:model-width}.}

\begin{figure*}[t]
\centering
\includegraphics[width=.6\linewidth]{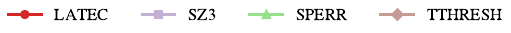}\\[0.1mm]
\includegraphics[width=.32\linewidth]{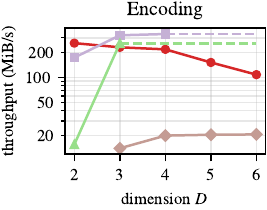}\hfill
\includegraphics[width=.32\linewidth]{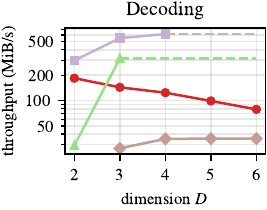}\hfill
\includegraphics[width=.32\linewidth]{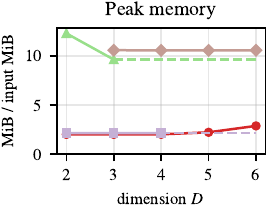}
\vspace{-6pt}
\caption{\rebuttalupdate{Throughput and peak memory growth (max of encoding and decoding) on 512\,MiB synthetic fields at RME $10^{-4}$ (\Cref{app:runtime-memory}). Dashed: beyond the codec's maximum dimension.}}
\label{fig:runtime-memory}
\vspace{-10pt}
\end{figure*}

\paragraph{Throughput.}
\rebuttalupdate{We time LATEC on one NVIDIA H100 GPU, on synthetic fields that SZ3 compresses to about 2 bits per point, at a relative error bound of $10^{-4}$, and the baselines on the same fields with 8 CPU threads of an AMD EPYC 9554 (\Cref{fig:runtime-memory}). LATEC encodes at 220-270\,MiB/s from 2-D to 4-D, 153\,MiB/s in 5-D and 108\,MiB/s in 6-D, and decodes at 187\,MiB/s in 2-D down to 78\,MiB/s in 6-D, with a fixed cost below 0.1\,s per call. Multithreaded SZ3 encodes 1.4-1.5 times faster than LATEC in 3-D and 4-D, and decodes 4-5 times faster, while in 2-D LATEC encodes 1.5 times faster than SZ3. SPERR is faster than LATEC in 3-D, but 16 times slower to encode and 6 times slower to decode in 2-D. TTHRESH, the only baseline that natively supports 5-D and 6-D tensors, encodes 5-7 times slower than LATEC there and decodes 2.2-2.8 times slower.}

\paragraph{Memory usage.}
\rebuttalupdate{Thanks to chunked execution, LATEC's memory grows linearly with the input, by 2.0-2.9\,MiB per MiB counting process RAM and GPU memory together, close to SZ3 (2.1), whereas SPERR and TTHRESH need 9.6-12.3 (\Cref{fig:runtime-memory}). For a 4\,GiB tensor, LATEC thus needs 8-12\,GiB, against 38-49\,GiB for SPERR and TTHRESH, and $64D$ times the input without chunking. Per-dimension regressions are given in \Cref{fig:runtime-memory-regressions} and \Cref{tab:latec-regressions}.}

\section{Conclusion}

\rebuttalupdate{Although LATEC needs a GPU and is slower than SZ3 from 3-D on, it suits the many workloads where data are compressed once and stored for long or read rarely, so that rate matters more than coding time.} Bitwise-reproducible decoding also ties streams to the platform that produced them, which deterministic integer inference would lift. Finally, LATEC's advantage is smaller at matched PSNR, where transform-based compressors remain strong; its gate could naturally be extended to include transform-based predictions, a promising direction for future work.

LATEC shows that one learned predictor, pretrained only on synthetic fields, can code scientific tensors of three to five dimensions in their native shapes at a lower rate than established error-bounded codecs, while strictly enforcing the pointwise bound these workflows require. It removes the two costs that kept learned compression out of these workflows: collecting a representative training corpus for each domain, and giving up the pointwise error guarantee. We hope this encourages general-purpose learned compressors for scientific data.

\subsection*{AI use}

Large language models assisted with code, literature search, drafting, and proofreading. Ideas, method, experiments, and figures are the authors' own, who verified all AI-assisted content.

\subsection*{Reproducibility}

LATEC's code, training scripts, and checkpoints (\url{https://github.com/etyl/LATEC}) and the Benchopt benchmark (\url{https://github.com/benchopt/benchmark_scidata_compression}) are public. All evaluation data are public except the Gysela field.

\subsection*{Acknowledgments}

This work was supported by the French National Research Agency (ANR) attributed to the ExaDoST project of the NumPEx PEPR program (ANR-22-EXNU-0004), and through the BenchArk project (ANR-24-IAS2-0003).

\bibliography{main}
\bibliographystyle{arxiv}

\clearpage \appendix
\counterwithin{figure}{section}
\counterwithin{table}{section}

\section{Causal schedule and predictor design}

\subsection{Causal multidimensional schedule}
\label[appendix]{app:causal-schedule}

\begin{figure}[t]
\centering
\includegraphics[width=.7\linewidth]{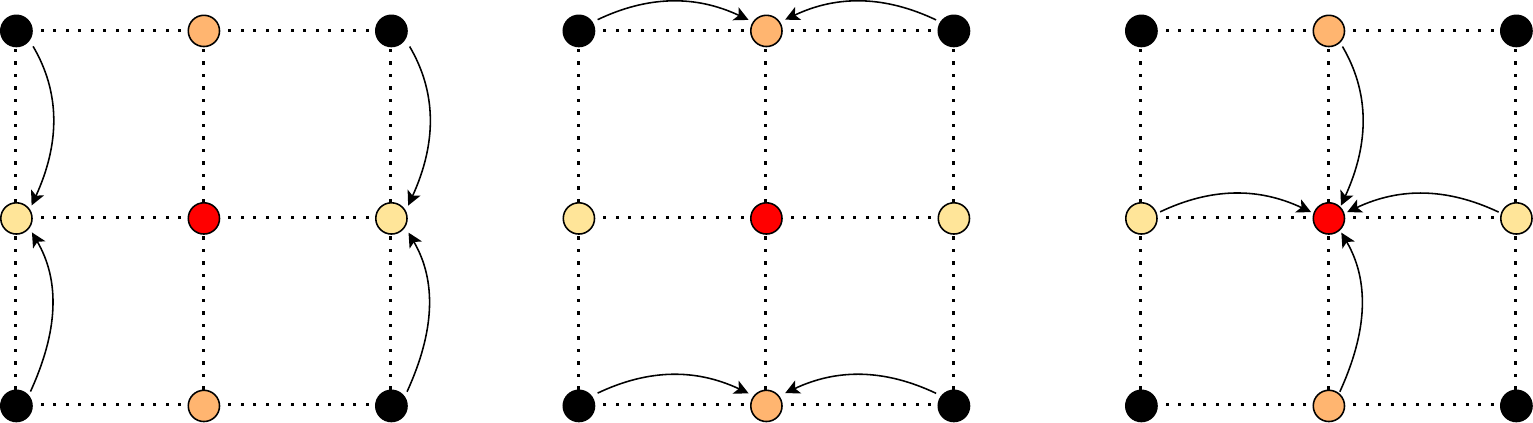}
\caption{The three sub-stages of one 2-D dyadic refinement. In 3-D there are five sub-stages: three for 1-D interpolation, one for 2-D, and one for 3-D.}
\label{fig:dyadic-schedule}
\end{figure}

The schedule has $K$ dyadic levels and stage strides $s_i=2^i$. Stage $K$ codes the anchor lattice, of stride $s_K=2^K$, directly against a zero predictor. Each refinement level $i=K-1,\ldots,0$ halves the stride. For each nonempty midpoint-axis set $S\subseteq\{1,\ldots,D\}$, a sub-stage comprises points whose offset $p_j-1$ from the first lattice index is an odd multiple of $s_i$ on axes in $S$ and a multiple of $2s_i$ elsewhere:
\begin{equation}
 \mathcal{M}_{i,S}=\left\{p:
 p_j-1\equiv s_i\pmod{2s_i}\;(j\in S),\
 p_j-1\equiv0\pmod{2s_i}\;(j\notin S)\right\}\setminus\mathcal{K}_{>i,S}.
 \label{eq:stage-mask}
\end{equation}
Here $\mathcal{K}_{>i,S}$ contains locations already decoded at coarser levels. The masks $\mathcal{M}_{i,S}$, one per nonempty $S$ at each level, form a disjoint partition into $1+K(2^D-1)$ sets. Sub-stages are processed by increasing $|S|$, and only the $D$ singleton sets ($|S|=1$) are kept separate by axis; sets with $|S|>1$ are merged by cardinality into a single sub-stage each, since same-cardinality sets never depend on one another. This groups the $2^D-1$ masks at each level into $2D-1$ sub-stages, for $1+K(2D-1)$ overall. Before a midpoint is predicted, its $\pm s_i$ neighbors along every axis in $S$ have already been reconstructed. \Cref{fig:dyadic-schedule} illustrates the resulting anchor-edge-center order in 2-D; the same rule generalizes to any number of dimensions.

This construction instantiates in any number of dimensions the dependency ordering used by multidimensional interpolation schemes such as HPEZ, which first interpolate points with 1-D context and then those with multidimensional context~\citep{liu2024hpez}. Simultaneous midpoints on every nonempty subset of axes become explicit sub-stages, providing decoder-reproducible causal geometry to the learned predictor.

The per-stage tolerance $\varepsilon_i=\varepsilon\rho^{i}$ is parameterized by its coarsest factor $r$ through $\rho=r^{1/(K-1)}$ for $K>1$, so the coarsest refinement stage ($i=K-1$) uses $r\varepsilon$ regardless of the number of refinement levels. This allocation affects rate, not the pointwise guarantee.

\subsection{Geometric context construction}
\label[appendix]{app:geometric-context}

Index a candidate line by $\ell$, and choose an oriented representative $u_\ell$ for its direction. If they exist, its nearest already reconstructed neighbors of $p$ in the $+u_\ell$ and $-u_\ell$ directions are $r^+_{p,\ell}$ and $r^-_{p,\ell}$, respectively. For axis $j$, the neighbor embeddings used in \Cref{eq:line-embedding} are $E^\pm_{\ell,j}=E_{r^\pm_{p,\ell},j}$, and their lattice distances are $\Delta^\pm_{p,\ell}=\lVert r^\pm_{p,\ell}-p\rVert_\infty\in\mathbb{N}$. Let $z^\pm_{\ell,j}$ be a learned lookup of the signed cosine $\pm u_{\ell,j}/\lVert u_\ell\rVert_2\in\{0,\pm1\}$. The context embeddings in \Cref{eq:line-embedding} collect the direction and distance features as $Z^\pm_{\ell,j}=[z^\pm_{\ell,j},\log_2\Delta^\pm_{p,\ell}]$ in the two-sided case. For a one-sided neighborhood, $\sigma\in\{+1,-1\}$ denotes the available side and $Z^\sigma_{\ell,j}=[z^\sigma_{\ell,j},\sigma,\log_2\Delta^\sigma_{p,\ell}]$. Missing neighbors and out-of-domain lines are masked before pooling.

\subsection{Aggregation-level tradeoff}
\label[appendix]{app:aggregation-level}

The axial GNN uses only axis-aligned lattice lines, corresponding to aggregation level $L_{\rm agg}=1$. We also considered $L_{\rm agg}=2$, which augments the $D$ axial lines with $D(D-1)$ two-axis diagonal lines. Each line contributes one embedding to every axis slot. At fixed embedding width, the $D$ axial lines therefore require $\mathcal{O}(D^2)$ embedding computations per query, while adding the diagonal lines raises this cost to $\mathcal{O}(D^3)$. Because the additional diagonal context yielded only marginal performance gains while making inference slower, all reported experiments use $L_{\rm agg}=1$.

\section{Deployed codec and implementation details}
\label[appendix]{app:entropy-rans}

\subsection{Chunk state and deployment settings}
\label[appendix]{app:chunk-state}

The encoder and decoder use the same state lifecycle. First, they reconstruct the global anchor lattice. For each chunk in raster order, they form its extended chunk and exclude its inherited low-coordinate faces from coding ownership. These one-point-wide faces contain adjacent values already reconstructed by earlier chunks. The codec reads their values from the global reconstruction buffer and includes the corresponding points in the new chunk's local stage sequence. When such a point $r$ is finalized, its embedding is rebuilt as $E_{r,j}=\phi_{\rm mix}(C_{r,j},\phi_{\rm val}(\widehat y_r))$, where $C_{r,j}$ contains only context rebuilt during the current chunk. No finalized embedding from the preceding chunk is retained.

Within the chunk, each stage finalizes the preceding stage's reconstructed points, computes the next predictions, and reconstructs the owned points from their residual symbols or escapes. Reconstructions are written to the global scalar buffer, including the chunk's owned high faces. After the last stage, the chunk is marked complete and its embedding field and cached stage context are released. A subsequent chunk reads its inherited faces from that buffer and rebuilds their embeddings during its own stage sequence. This state reset limits the learned context carried across a seam, but keeps both sides synchronized without storing a global neural field.

Each chunk is one anchor cell, with edge equal to the anchor stride $s_K=2^K$ along every axis, clipped at the domain boundary. The one-point shared faces add to this nominal volume. For the aggregation-level-one model, automatic refinement depths are 7, 5, and 4 in 3-D, 4-D, and 5-D, respectively, giving nominal chunk volumes of $128^3=2^{21}$, $32^4=2^{20}$, and $16^5=2^{20}$ points. These depths are chosen large enough to give each chunk ample coarse context, while keeping the per-chunk memory footprint moderate. The depth is capped by $\max(1,\lfloor\log_2 n_{\max}\rfloor-1)$ on small tensors, where $n_{\max}=\max_j n_j$. The default coarse error factor is $r=0.9$.

\subsection{Bounded reconstruction and escapes}
\label[appendix]{app:error-control}
For a finite radius $R$, an in-range quantization index from \Cref{eq:quant-index} ($|q_p|<R$) is sent as $c_p=q_p+R\in\{1,\ldots,2R-1\}$; $c_p=0$ is an escape symbol. The encoder emits this escape symbol, followed by the original normalized float32 value in scan order, if the index is out of range or if verification of \Cref{eq:reconstruction} fails. Thus every non-escape value satisfies $|y_p-\widehat y_p|\leq\varepsilon_i$ by explicit verification, and every escape value is reproduced exactly in normalized float32. Normalization and the stored reconstructed state use float32. Residual-index calculation and reconstruction accumulation use float64 before the candidate is cast to float32 and verified. Decoder reconstruction uses the same operations. Verification is performed in normalized coordinates. In exact arithmetic, inverse normalization multiplies this error by the original data range, so the normalized bound is the range-relative bound in original units. Normalization and inverse normalization introduce floating-point rounding, and the benchmark evaluates the final output error in float64.

This is the same separation of concerns used by predictive error-bounded compressors: prediction improves rate, whereas quantization and an escape path establish the bound~\citep{di2016fastsz,liang2022sz3}. It also appears in learned scientific codecs: AE-SZ combines a blockwise autoencoder with SZ-style error-controlled coding~\citep{liu2021aesZ}, \citet{lee2022errorbounded} apply an error-bounded residual post-process after learned reduction, and SRN-SZ alternates neural super-resolution with quantized correction offsets~\citep{liu2023srnsz}. In each case, the learned model improves the reconstruction estimate, but the explicit residual or correction representation supplies the pointwise guarantee.

\subsection{Gate selection and transmission}
\label[appendix]{app:gate}
This section details the gate of \Cref{eq:gate}: its classical interpolators, the encoder search that selects one gate per chunk-stage, and how the selection is passed on to the decoder.

\textbf{Classical interpolators.} At stage $i$, a point $p$ is the midpoint of its neighbors at distance $s_i$ along each of its midpoint axes, the axes in its set $S$ of \Cref{eq:stage-mask}. Along one such axis, linear interpolation averages the two neighbors at $\pm s_i$, and cubic interpolation uses the four-tap stencil $(-1,9,9,-1)/16$ on the neighbors at $\pm s_i,\pm3s_i$. Near a line end, where the $\pm3s_i$ neighbor is missing, the cubic stencil falls back to a quadratic one, and beyond the last available neighbor both interpolators extrapolate. The three interpolators $a$ are
\begin{itemize}
 \item \emph{cubic}: the cubic interpolations averaged over all midpoint axes in 3-D and below, and the cubic interpolation along the first midpoint axis in 4-D and above, where averaging many estimates was found to be less accurate.
 \item \emph{first-axis linear}: linear interpolation along the first midpoint axis.
 \item \emph{last-axis linear}: linear interpolation along the last midpoint axis.
\end{itemize}
They read only reconstructed values, so encoder and decoder compute identical candidates.

\textbf{Gated mean and scale.} The gate of a chunk-stage is a tuple $(a,d,T,s)$ that consists of a classical interpolator $a$, a side $d\in\{\mathrm{low},\mathrm{high}\}$, a threshold exponent $T\in\{2,\ldots,13\}$, and a scale shift $s\in\{0,2,4,6\}$. It selects the points $g_p=1$ of \Cref{eq:gate}: those with $\beta_p<\varepsilon_i2^T$ for the low side, and the others for the high side. At these points, the gated mean $\widehat\mu_p$ is the classical interpolation and the gated scale is $\widehat\beta_p=2^{-s}\beta_p$. Elsewhere, both the GNN mean and scale are kept. Since $\beta_p\in[\varepsilon_i/16,4096\varepsilon_i]$, the largest exponent $T=13$ lies above every scale, so its low side replaces the GNN mean on the whole chunk-stage, whereas smaller exponents replace it only on part of the chunk-stage. The shift is shared by all selected points rather than chosen per point, so $\widehat\beta_p$ keeps the pointwise variation of $\beta_p$. Both $\widehat\mu_p$ and $\widehat\beta_p$ are then used exactly as the ungated outputs: the residual $y_p-\widehat\mu_p$ is quantized under $\varepsilon_i$, and $\widehat\beta_p$ selects its probability table (\Cref{app:entropy}).

\textbf{Encoder search.} For every nonempty chunk-stage, the encoder evaluates all $3\times2\times12\times4=288$ tuples and the GNN alone, and keeps the one with the smallest estimated code length. The estimated length of a stage is the sum over its points of the cost of the residual index under the Laplace model of \Cref{eq:laplace-mass}, after snapping $\widehat\beta_p$ to the 64-level scale grid used for coding. A regular index costs its negative log probability, capped at 24 bits, and an out-of-range index costs 56 bits, a 24-bit escape marker plus its raw float32 value. The GNN alone is the baseline, and the gate is disabled unless some tuple is strictly cheaper. The search is exhaustive yet inexpensive: the points are binned by the interval between consecutive thresholds $\varepsilon_i2^T$ containing their scale, and cumulative sums of per-bin costs give the cost of every threshold and side at once. The search compares residuals to the source values, which only the encoder holds, but it selects among predictors and therefore affects rate only. Quantization and verification enforce the error bound for any choice.

\textbf{Transmission to the decoder.} The selected tuple is packed into one 16-bit word: two bits for the interpolator, with zero meaning that the gate is off, one bit for the side, four bits for $T$, and four bits for $s$. The words of all chunk-stages are compressed with Zstandard at level 19 and stored base64-encoded in the stream header. At each chunk-stage, the decoder reads the word, computes the selected interpolation from reconstructed values, and applies \Cref{eq:gate} to the scale $\beta_p$ it recomputes from the GNN. It thus rebuilds the same mask, gated mean, and gated scale as the encoder without repeating the search, and no per-point flag is transmitted. Predictor-selection percentages in \Cref{fig:gate-selection} count the points selected by these chunk-stage rules.

\subsection{Entropy model and rANS coding}
\label[appendix]{app:entropy}

The error-control procedure determines the symbols; the entropy model affects only their code length. For each point, the gate supplies the adjusted Laplace scale $\widehat\beta_p$. We round it to one of 64 logarithmically spaced values in $[\varepsilon_i/16,4096\varepsilon_i]$, \rebuttalupdate{denoted $\lfloor\widehat\beta_p\rceil$}, allowing encoder and decoder to select the same probability table without transmitting a floating-point scale per value. \rebuttalupdate{The grid spans a factor of $2^{16}$ in 63 steps, so consecutive values differ by the factor $\eta=2^{16/63}\approx1.19$. Rounding therefore changes a scale by at most $\sqrt\eta\approx1.09$, which, for a continuous Laplace model, adds at most about $0.006$ bits per point, against typical rates of 2 to 8 bits per point. At the low end of the range, almost all of the probability mass falls in the zero bin of width $2\varepsilon_i$, so finer scales would give practically the same table; the high end must stay far above $\varepsilon_i$ because, at tight bounds, prediction errors are orders of magnitude larger than $\varepsilon_i$. The coded scale $\widetilde\beta_p=\eta^{\kappa}\lfloor\widehat\beta_p\rceil$ then applies the per-chunk-stage shift $\kappa$ of \Cref{eq:scale-shift}, described below.} The probability assigned to the point's index $q_p$ is the mass of its quantization bin,
\begin{equation}
 P(q_p\mid\widetilde\beta_p)=F_{\rm Lap}(2\varepsilon_iq_p+\varepsilon_i;0,\widetilde\beta_p)
 -F_{\rm Lap}(2\varepsilon_iq_p-\varepsilon_i;0,\widetilde\beta_p).
 \label{eq:laplace-mass}
\end{equation}
At each stage, the encoder records the smallest power-of-two alphabet that covers the observed in-range indices and a selected core-tail Laplace-mixture model. From these stage-level parameters\rebuttalupdate{, the shift $\kappa$ described below,} and the rounded per-point scale, the decoder regenerates the same normalized integer CDF. rANS then encodes the residual or escape symbols using that CDF~\citep{duda2013ans}\rebuttalupdate{, on interleaved GPU lanes whose states carry over from one stage to the next}. The float32 values for escape symbols are stored immediately after the rANS payload and consumed in symbol order. The complete payload is finally framed with Zstandard lossless compression.

\rebuttalupdate{\textbf{Per-chunk-stage scale shift.} The learned scale can be systematically over- or under-confident on a chunk-stage, for instance on fields far from the pretraining distribution. Each chunk-stage may therefore shift all its scale-grid levels by one signed integer $\kappa$: a point with grid level $l_p$ is coded under level $\min(\max(l_p+\kappa,0),63)$, which multiplies its scale by $\eta^{\kappa}$ (\Cref{eq:scale-shift}). Every point keeps its own predicted scale, so the shift removes a uniform miscalibration without discarding the pointwise variation of $\widetilde\beta_p$. The encoder selects $\kappa$, with $|\kappa|\le21$, so the shift can multiply or divide every scale of the chunk-stage by up to $\eta^{21}=2^{16/3}\approx40$. It selects $\kappa$ jointly with the mixture model to minimize the estimated code length of the chunk-stage, first over every third shift and then over the neighbors of the best one, and transmits it in six bits of a 16-bit stage descriptor. The shift acts only on the entropy model and is distinct from the gate shift $s$ of \Cref{app:gate}, which applies only to gated points.}

The quantizer radius is $R=2^{15}$, and integer CDFs use precision 24, assigning at least one count to each symbol. The stage model is selected from a pure Laplace model and 20 core-tail mixtures. A mixture combines scale-grid level $l$ with level $\min(l+\delta,63)$, using $\delta\in\{4,8,16,24\}$ and tail weight $w\in\{0.02,0.05,0.12,0.25,0.45\}$. Selection minimizes estimated negative log probability on \rebuttalupdate{up to 384 evenly strided blocks of 128 consecutive points (about 50,000 points), or all points in smaller stages. Every mixture is available at every alphabet size.} The payload uses Zstandard level 9.

\section{Training procedure and hyperparameters}
\label[appendix]{app:training-details}

\subsection{Objective and simulated reconstruction noise}
Training uses teacher-forced reconstructed values to expose the predictor to imperfect context, as stated in \Cref{sec:training-stream}. A separate $\varepsilon$ is sampled for each field, log-uniformly from $10^{-6}$ to $10^{-2}$. The same sampled tolerance conditions every training stage and sets the perturbation amplitude. This is a surrogate for deployment, which can assign tighter tolerances to coarse stages; training does not simulate that allocation or the quantized feedback exactly. Evaluation additionally includes $10^{-1}$, outside the training range.

Write $r_p=y_p-\mu_p$ for the continuous residual, so that the summand of \Cref{eq:training-objective} integrates the Laplace density over $[r_p-\varepsilon,r_p+\varepsilon]$. Neither the residual index nor the scale-table index is rounded, so gradients reach both $\mu_p$ and $\beta_p$ directly; the loss is a differentiable rate surrogate rather than the exact deployed stream length.

For large residuals, our implementation softens the Laplace tail to avoid gradients proportional to an extremely small inverse scale. When $|r_p|\geq\varepsilon$, define $t_p=(|r_p|-\varepsilon)/\beta_p$ and $t_0=32\ln 2$, and replace $t_p$ by
\begin{equation}
 g(t_p)=\begin{cases}
 t_p,&t_p\leq t_0,\\
 t_0+\ln(1+t_p-t_0),&t_p>t_0.
 \end{cases}
\end{equation}
The optimized objective is then the stabilized form of \Cref{eq:training-objective},
\begin{equation}
 \widetilde{\mathcal L}=\frac{1}{|\mathcal H|}\sum_{p\in\mathcal H}
 \begin{cases}
 -\log_2\!\left[F_{\rm Lap}(r_p+\varepsilon;0,\beta_p)-F_{\rm Lap}(r_p-\varepsilon;0,\beta_p)\right],
 &|r_p|<\varepsilon,\\[4pt]
 \dfrac{\ln 2+g(t_p)-\ln\!\left(1-\exp(-2\varepsilon/\beta_p)\right)}{\ln 2},
 &|r_p|\geq\varepsilon,
 \end{cases}
 \label{eq:training-loss}
\end{equation}
where the first branch floors the probability at $10^{-30}$ for numerical evaluation. The tail continuation preserves a mean gradient for extreme errors while reducing its magnitude. It approximates the limited usefulness of further rate penalties where deployment can escape to a raw value. Differentiation passes through the learned context and finalized embeddings within a stage sequence. Teacher-forced values do not depend on earlier mean predictions, so gradients do not pass through a simulated quantizer. Chunked training uses perturbed boundary reconstructions and discards the preceding chunk's embeddings, exposing the predictor to the deployed absence of cross-seam learned state. The classical gate, scale-grid rounding, and stage-level entropy-model selection are deployment operations and are not optimized in this training loop.

\subsection{Synthetic pretraining}
We generate fresh fields at every optimizer step using the distribution described in \Cref{sec:training-stream}. Each step contains 16 fields of shape $128^2$ and four fields of shape $16^4$, so the 2-D and 4-D fields each contribute 262,144 scalar values and receive equal weight in the loss. For the 2-D fields, we sample the anchor stride uniformly from $\{8,16,32\}$ (respectively three, four, or five refinement levels); half of these steps randomly split the field into two chunks and randomize both the split axis and chunk traversal order, exposing the model to the deployed halo exchange. The 4-D fields use stride 8 and three levels.

The construction types of \Cref{tab:synthetic-realizations} are drawn with the continuous parameters resampled within each type, as follows.

\begin{table}[t]
\centering
\small
\setlength{\tabcolsep}{3pt}
\caption{Possible synthetic-field constructions. Each realization combines one spectrum, one value warp, and one discontinuity choice. Here $f(\mathbf{x})$ denotes the current field value at grid coordinate $\mathbf{x}$. Parameter distributions are given in the accompanying text.}
\label{tab:synthetic-realizations}
\begin{tabular}{@{}p{0.17\linewidth}p{0.22\linewidth}p{0.55\linewidth}@{}}
\toprule
Group & Realization & Formula or operation \\
\midrule
Spectrum
& Smooth
& Two-scale anisotropic Gaussian filtering; see \Cref{eq:synthetic-field-sum,eq:synthetic-correlation,eq:synthetic-smooth-filter}. \\
& Turbulent
& Radial power-law filtering followed by repeated smooth displacement; see \Cref{eq:synthetic-field-sum,eq:synthetic-turbulent-filter,eq:synthetic-warp}. \\
\midrule
Value warp
& Identity
& $f(\mathbf{x})\mapsto f(\mathbf{x})$. \\
& Signed power
& $f(\mathbf{x})\mapsto\operatorname{sign}(f(\mathbf{x}))|f(\mathbf{x})|^\gamma$. \\
& Two-phase
& $f(\mathbf{x})\mapsto\tanh(\gamma f(\mathbf{x}))$. \\
\midrule
Jumps
& None
& $f(\mathbf{x})\mapsto f(\mathbf{x})$. \\
& One to three hyperplanes
& $f(\mathbf{x})\mapsto f(\mathbf{x})+\sum_{c=1}^{C}\Delta_c\mathbf 1\{\boldsymbol{\nu}_c^\top\mathbf{x}>t_c\}$. \\
\bottomrule
\end{tabular}
\end{table}

For a field on an $n_1\times\cdots\times n_D$ grid, let $\mathbf{k}=(k_1,\ldots,k_D)$ contain the discrete FFT frequencies in cycles per grid cell. We independently sample two standard-normal white-noise fields $\xi_1$ and $\xi_2$ and mirror-extend each axis to impose reflecting boundaries. Let $\mathcal F$ denote the $D$-dimensional FFT. Given amplitude filters $G_1$ and $G_2$, the initial field is
\begin{equation}
 f^{(0)}(\mathbf{x})=
 \mathcal F^{-1}\!\left[G_1\mathcal F\xi_1\right](\mathbf{x})
 +\frac{1}{2}\mathcal F^{-1}\!\left[G_2\mathcal F\xi_2\right](\mathbf{x}).
 \label{eq:synthetic-field-sum}
\end{equation}
The realization is smooth with probability $0.5$ and turbulent otherwise.

For a smooth realization, we draw a base width $b$ log-uniformly from $[1,32]$ cells in 2-D or $[0.5,8]$ cells in 4-D, draw an anisotropy spread $a\sim\mathcal U[0,1]$, and set
\begin{equation}
 \sigma_j=\max\!\left(0.3,b\exp(\delta_j)\right),\qquad
 \delta_j\sim\mathcal N(0,a^2).
 \label{eq:synthetic-correlation}
\end{equation}
The two Gaussian amplitude filters are
\begin{equation}
 H_s(\mathbf{k})=\exp\!\left(-2\pi^2\sum_j(s\sigma_j)^2k_j^2\right),
 \qquad s\in\{1,2\}.
 \label{eq:synthetic-smooth-filter}
\end{equation}
We use $G_1=H_1$ and $G_2=H_2$ in \Cref{eq:synthetic-field-sum}. The configured correlation inputs are $[2,32]$ and $[1,8]$ for 2-D and 4-D, respectively. The implementation draws $b$ from half the lower input through the upper input, which gives the intervals above.

For a turbulent realization, we draw $p\sim\mathcal U[0.7,1.6]$ and use the radial amplitude filter
\begin{equation}
 H_{\rm turb}(\mathbf{k})=
 \begin{cases}
 (\lVert\mathbf{k}\rVert_2^2+10^{-8})^{-p/2},&\lVert\mathbf{k}\rVert_2>0,\\
 0,&\lVert\mathbf{k}\rVert_2=0.
 \end{cases}
 \label{eq:synthetic-turbulent-filter}
\end{equation}
Here $p$ is the Fourier-amplitude exponent. We use $G_1=G_2=H_{\rm turb}$ in \Cref{eq:synthetic-field-sum}.

The turbulent field is then advected by a smooth random displacement. Let $m=\min_j n_j$. For each component $j$, we Gaussian-filter an independent white-noise scalar field with isotropic width $\ell\sim\mathcal U[0.18m,0.40m]$, producing a scalar field $v_j(\mathbf{x})$ that we normalize to unit spatial standard deviation. Together, these components form the vector field $\mathbf{v}(\mathbf{x})=(v_1(\mathbf{x}),\ldots,v_D(\mathbf{x}))$. A single $\ell$ is shared by the fields in a generated batch, while their component noise is independent. Each turbulent field draws $A\sim\mathcal U[0.04m,0.10m]$ and sets $\mathbf{u}(\mathbf{x})=A\mathbf{v}(\mathbf{x})$. Starting from $f^{(0)}$, the generator applies the same frozen displacement $\mathbf{u}$ three times,
\begin{equation}
 f^{(r+1)}(\mathbf{x})=f^{(r)}(\mathbf{x}+\mathbf{u}(\mathbf{x})),
 \qquad r\in\{0,1,2\},
 \label{eq:synthetic-warp}
\end{equation}
using multilinear interpolation and reflecting boundaries. Smooth fields are not spatially warped.

\begin{figure*}[t]
\centering
\begin{subfigure}[t]{0.188\textwidth}
  \centering
  \includegraphics[width=\linewidth]{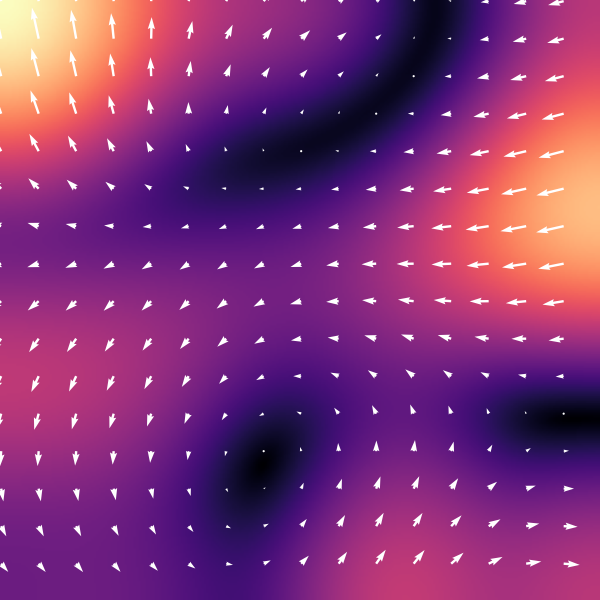}
  \caption{Displacement $\mathbf{u}$}
\end{subfigure}\hfill
\begin{subfigure}[t]{0.188\textwidth}
  \centering
  \includegraphics[width=\linewidth]{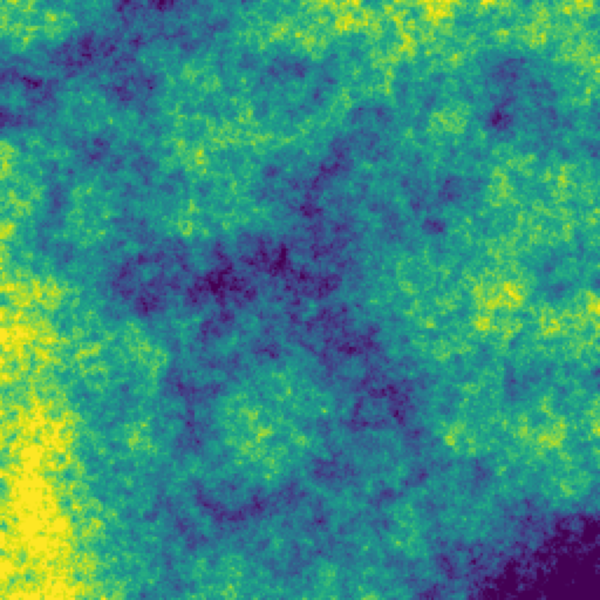}
  \caption{Initial field $f^{(0)}$}
\end{subfigure}\hfill
\begin{subfigure}[t]{0.188\textwidth}
  \centering
  \includegraphics[width=\linewidth]{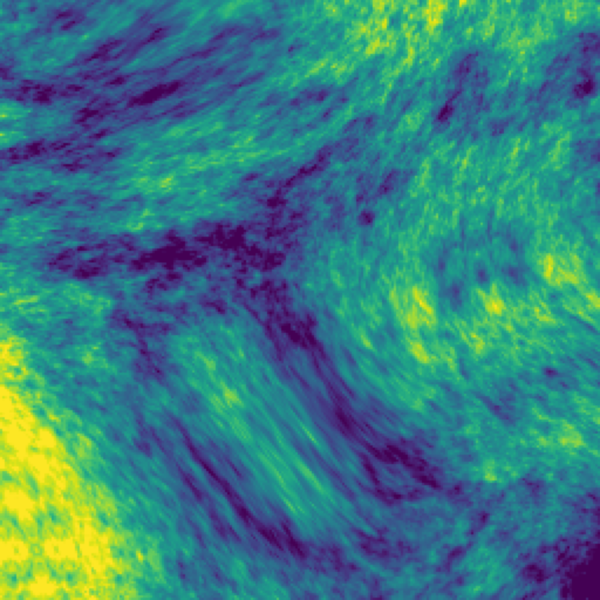}
  \caption{First warp $f^{(1)}$}
\end{subfigure}\hfill
\begin{subfigure}[t]{0.188\textwidth}
  \centering
  \includegraphics[width=\linewidth]{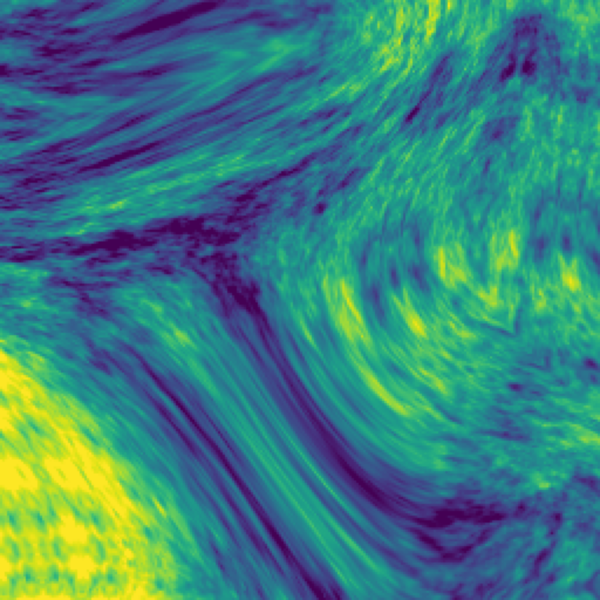}
  \caption{Second warp $f^{(2)}$}
\end{subfigure}\hfill
\begin{subfigure}[t]{0.188\textwidth}
  \centering
  \includegraphics[width=\linewidth]{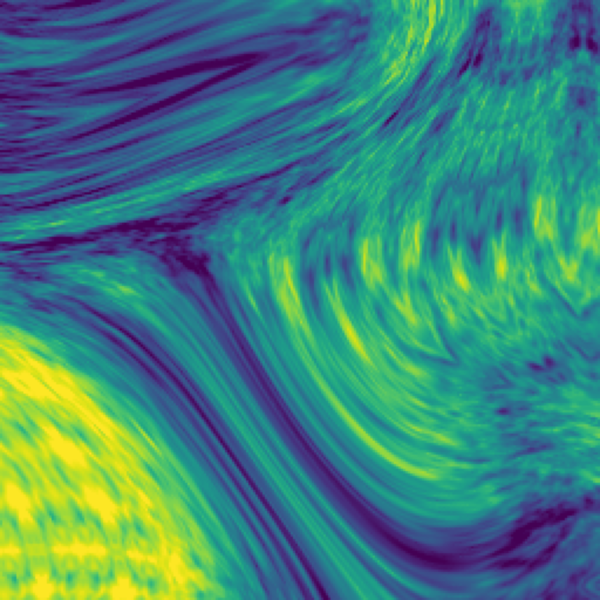}
  \caption{Third warp $f^{(3)}$}
\end{subfigure}
\caption{Construction of a turbulent synthetic field in 2-D. A smooth random displacement $\mathbf{u}$, shown by arrows over its magnitude, repeatedly advects the initial power-law field $f^{(0)}$. Reusing the same displacement for three warps progressively bends the multiscale structure into coherent filaments.}
\label{fig:turbulent-generation}
\end{figure*}

\Cref{fig:turbulent-generation} visualizes this construction in 2-D. The same frozen displacement progressively bends the initial multiscale texture into filamentary structures.

Finally, each field $f$ is standardized before applying one of the three value warps in \Cref{tab:synthetic-realizations}, chosen with equal probability. The signed-power exponent is $\gamma\sim\mathcal U[0.4,3]$, while the two-phase exponent is $\gamma\sim\mathcal U[1.5,4]$. With probability $0.25$, we draw $C$ uniformly from $\{1,2,3\}$ and add the hyperplane jumps shown in the table. Each normal is $\boldsymbol{\nu}_c=\mathbf{g}_c/\lVert\mathbf{g}_c\rVert_2$ for $\mathbf{g}_c\sim\mathcal N(0,I_D)$. Its threshold $t_c$ is the $q_c$-quantile of $\boldsymbol{\nu}_c^\top\mathbf{x}$ over the grid, where $q_c\sim\mathcal U[0.25,0.75]$. Its signed magnitude is $\Delta_c=\epsilon_cJ_cs_f$, where $\epsilon_c$ is uniform on $\{-1,1\}$, $J_c\sim\mathcal U[0.4,1.5]$, and $s_f$ is the standard deviation of $f$ before adding any jumps. The result is mapped to $[0,1]$ using its 1st and 99th percentiles and clipped to that interval.

The reported pretrained checkpoint has embedding width $d=64$, aggregation level $L_{\rm agg}=1$, and 102,710 trainable parameters. We train it for 50,000 optimizer steps with AdamW, learning rate $5\times10^{-4}$, weight decay $10^{-3}$, and the default $(\beta_1,\beta_2)=(0.9,0.999)$. The learning rate increases linearly during the first 1,000 steps, remains constant through 60\% of training, and then decreases linearly to zero over the final 40\%. We maintain an exponential moving average of the weights with decay $0.999$ and save these averaged weights. Pretraining runs on a single NVIDIA H100 and takes approximately 4 hours.

\begin{figure}[t]
\centering
\includegraphics[width=\linewidth]{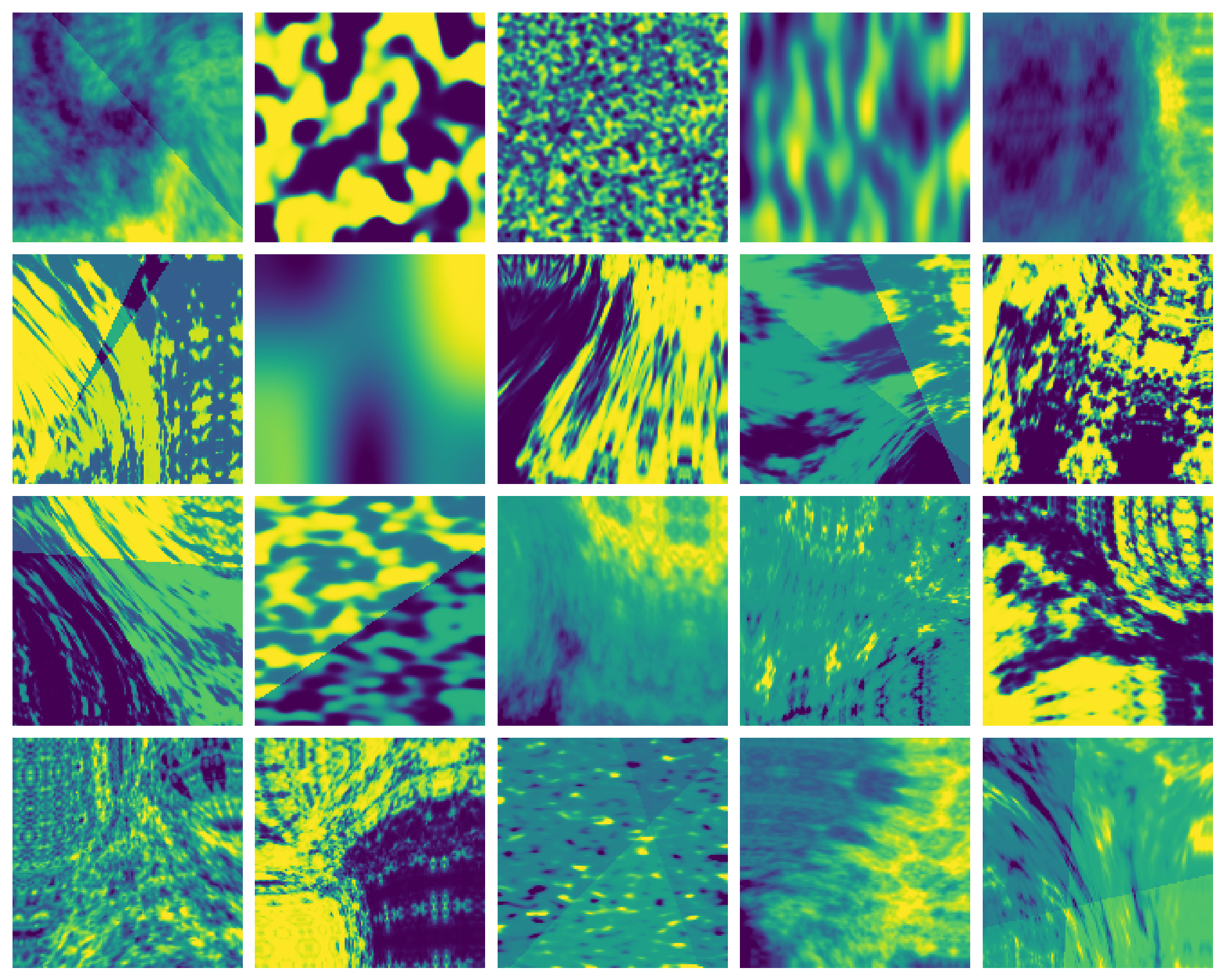}
\caption{Examples of $128\times128$ 2-D fields drawn from the synthetic training distribution. The realizations span smooth and multiscale correlation, anisotropy, turbulent filaments, and sharp hyperplane jumps.}
\label{fig:synthetic-2d-samples}
\end{figure}

\subsection{Dataset-specific fine-tuning}
Fine-tuning starts from the same pretrained $d=64$, $L_{\rm agg}=1$ checkpoint. A batch is formed from independently located, min-max-normalized hypercubic crops. For 3-D tensors, crops have side 32 and the schedule level is sampled uniformly from $\{3,4,5\}$; for 4-D tensors, crops have side 16 and levels are sampled from $\{2,3,4\}$. One schedule depth is drawn per optimizer step. GyroSwin's 5-D distribution-function fields use crops capped at 16 cells per axis (shorter axes are left whole) and levels $\{2,3,4\}$. \rebuttalupdate{For Gysela, which has no separate run, we fine-tune for 20,000 steps, about 16 hours on one H100, on eight crops that are spatially disjoint from the test crop.} These depth ranges deliberately train several rollout lengths instead of specializing the weights to one hierarchy. We use 10,000 steps, batch size 16 for The Well and batch size 1 for GyroSwin, AdamW with learning rate $5\times10^{-5}$ and weight decay $10^{-3}$, a 500-step linear warmup, linear decay over the final 40\%, and EMA decay $0.999$. The error-bound sampling and rate loss are unchanged from pretraining. Fine-tuning runs on a single NVIDIA H100 and takes approximately 1 hour for 3-D, 2 hours for 4-D and 4 hours for GyroSwin. All reported \rebuttalupdate{rate} evaluations run on a single NVIDIA V100\rebuttalupdate{; runtime and LATEC memory are measured on an H100 (\Cref{app:runtime-memory})}.

\subsection{\rebuttalupdate{Per-input training}}
\label[appendix]{app:per-input}
\rebuttalupdate{Neural-field and NeurLZ-style codecs~\citep{jia2024neurlz} adapt the network to each input. To test this setting, we fine-tune the pretrained checkpoint for $k$ steps on the input tensor itself and store the resulting weights as an XOR delta to the pretrained weights. The delta takes 81-141\,KiB and is counted in the stream. \Cref{tab:per-input} reports the change in total size after $k=1000$ steps. Per-input training saves 5.9-7.2\% on Gysela and GyroSwin and 1.0-1.8\% on MHD-64, at the cost of 3-29 minutes of H100 training, compared with 1.6-20 seconds of encoding. The effect is not monotone in $k$: after 50-200 steps the stream is up to 20\% larger than with the pretrained weights. The weight delta costs 0.005-0.007\,BPP on Gysela but 0.05-0.09\,BPP on the 11-17M-point fields, where it cancels about half of the payload gain. Fine-tuning once per dataset gives comparable or larger savings with no per-input cost, so per-input training is worthwhile only for large, long-lived tensors.}

\begin{table}[h]
\centering
\footnotesize
\caption{\rebuttalupdate{Per-input training: change in total BPP (\%; lower is better) after $k=1000$ steps on the input, relative to the pretrained checkpoint, with the weight delta included in the stream. Compression time excludes per-input training.}}
\label{tab:per-input}
\begin{tabular}{lrrrrr}
\toprule
Field & Points & RME $10^{-3}$ & RME $10^{-4}$ & Training time & Compression time \\
\midrule
Gysela & 134M & $-7.2$ & $-5.9$ & 29\,min & 20\,s \\
GyroSwin ($t=100$) & 11M & $-6.5$ & $-5.9$ & 15\,min & 1.6\,s \\
MHD-64 & 17M & $-1.8$ & $-1.0$ & 2.7\,min & 2.6\,s \\
\bottomrule
\end{tabular}
\end{table}

\begin{table*}[t]
\centering
\footnotesize
\caption{Effect of synthetic pretraining dimensionality on the average relative BPP difference (\%; lower is better) at the same relative maximum error (RME) target, relative to SZ3. The dimensions in parentheses specify the dimensions of the synthetic fields used for pretraining, not the dimensionality supported at inference. Negative values denote a BPP saving. Blue, green, and orange column backgrounds denote 3-D, 4-D, and 5-D evaluation datasets, respectively. See \Cref{app:rate-gains} for the calculation. Bold indicates the best method on each dataset, and underlined indicates the second-best.}
\label{tab:latec-pretraining}
\vspace{-4pt}
\newcommand{\rotateheader}[1]{\makebox[0pt][c]{\rotatebox[origin=lb]{90}{\strut #1}}}
\begin{tblr}{
  colspec = {l*{13}{r}},
  colsep = 1.5pt,
  row{1} = {halign=c,valign=b,belowsep=2pt,cmd=\rotateheader},
  cell{1}{1} = {cmd={},halign=l},
  column{2-6} = {bg=blue!8},
  column{7-11} = {bg=green!10},
  column{12-14} = {bg=orange!12},
  hline{2} = {solid},
}
Method & \shortstack[l]{Euler-Multi-\\Quadrants} & Gray-Scott & \shortstack[l]{Hurricane\\ISABEL} & Miranda & S3D & \shortstack[l]{Convective-\\Envelope} & \shortstack[l]{Gravity-\\Cooling} & MHD-64 & RT-Instability & \shortstack[l]{Supernova-\\Explosion} & GyroSwin & Gysela & Synthetic 5D \\
SZ3 & \textbf{0} & 0 & 0 & 0 & 0 & \underline{0} & 0 & 0 & 0 & 0 & \textbf{0} & 0 & 0 \\
LATEC (2D) & 28.2 & 9.4 & -6.1 & -3.3 & -8.7 & 41.4 & -2.2 & 5.1 & -13.4 & -4.7 & 11.1 & 23.9 & -11.4 \\
LATEC (2D+3D) & \underline{16.9} & \textbf{-6.8} & \underline{-10.9} & \underline{-10.8} & \underline{-13.0} & \textbf{-2.2} & \textbf{-11.4} & \underline{-18.8} & \textbf{-36.3} & \underline{-9.7} & 3.0 & \textbf{-9.9} & \underline{-44.9} \\
LATEC (2D+4D) & 18.8 & \underline{-4.7} & \textbf{-14.8} & \textbf{-11.4} & \textbf{-14.5} & 1.9 & \underline{-3.4} & \textbf{-19.0} & \underline{-35.9} & \textbf{-9.9} & \underline{1.9} & \underline{-3.2} & \textbf{-47.3} \\
\end{tblr}
\vspace{-8pt}
\end{table*}

\subsection{Effect of pretraining dimensionality}
Adding either 3-D or 4-D synthetic fields to the 2-D pretraining distribution improves the rate of the pretrained checkpoint on every evaluation dataset. The 2-D+4-D checkpoint yields a lower BPP difference than the 2-D+3-D checkpoint on seven datasets, while the 2-D+3-D checkpoint is better on six. Neither checkpoint dominates simply by matching the evaluation dimensionality. Overall, the 2-D+3-D and 2-D+4-D checkpoints save BPP relative to SZ3 on eleven and ten datasets, respectively, compared with seven for 2-D-only pretraining. These results show that exposure to higher-dimensional synthetic structure is important for transfer, while the precise additional pretraining dimension has a smaller and dataset-dependent effect.

\section{Evaluation protocol}

\subsection{Evaluation metrics}
\label[appendix]{app:metrics}

For a field $\mathbf{x}$ with $N$ points and reconstruction $\widehat{\mathbf{x}}$,
\begin{equation}
  \operatorname{RME} = \frac{\lVert\mathbf{x}-\widehat{\mathbf{x}}\rVert_\infty}
       {\max_p x_p-\min_p x_p},
  \qquad
  \operatorname{PSNR} = 10\log_{10}\!\left(
    \frac{(\max_p x_p-\min_p x_p)^2}
         {\frac{1}{N} \lVert\mathbf{x}-\widehat{\mathbf{x}}\rVert_2^2}
  \right).
\end{equation}
The range-relative maximum error directly tests the pointwise guarantee: a requested normalized bound $\varepsilon$ corresponds to a horizontal target at $\varepsilon$ in the curves of \Cref{fig:rate-distortion}.

\subsection{Benchmark datasets}

\begin{table}[h]
\centering
\caption{The 13 fields used in the benchmark, sorted by size.}
\label{tab:datasets}
\begin{tabular}{lllr}
\toprule
Dataset & Field & Shape & Size (MiB) \\
\midrule
Gysela & density & $64\times64\times64\times64\times64$ & 4096 \\
Convective-Envelope & density & $64\times256\times128\times256$ & 2048 \\
RT-Instability & density & $64\times128\times128\times128$ & 512 \\
S3D & 0 & $500\times500\times500$ & 477 \\
Supernova-Explosion & density & $32\times128\times128\times128$ & 256 \\
Miranda & density & $256\times384\times384$ & 144 \\
Synthetic 5D & N/A & $32\times32\times32\times32\times32$ & 128 \\
Hurricane ISABEL & CLOUDf48 & $100\times500\times500$ & 95 \\
Euler-Multi-Quadrants & density & $64\times512\times512$ & 64 \\
MHD-64 & density & $64\times64\times64\times64$ & 64 \\
GyroSwin & density ($t$=100) & $32\times8\times16\times85\times32$ & 43 \\
Gravity-Cooling & density & $32\times64\times64\times64$ & 32 \\
Gray-Scott & A (gliders) & $256\times128\times128$ & 16 \\
\bottomrule
\end{tabular}
\end{table}

\subsection{Synthetic 5D dataset and benchmark adapters}
\label[appendix]{app:synthetic-5d}

Synthetic 5D extends the smooth Gaussian-field family used in pretraining to five axes. The benchmark draws float32 standard Gaussian noise on a $32^5$ grid with NumPy generator seed zero. It samples a base correlation length log-uniformly from $[1,8]$, an anisotropy spread $u\sim\mathcal U[0,0.8]$, and independent $z_j\sim\mathcal N(0,u^2)$. The smoothing widths are $\sigma_j=\max(0.5,\sigma_{\rm base}\exp(z_j))$. A separable Gaussian filter uses reflective boundaries and truncates at three standard deviations. The field is then centered, divided by its standard deviation, multiplied by a factor drawn uniformly from $[1,3]$, transformed with $\tanh$, and min-max normalized to $[0,1]$. This benchmark realization shares smoothness, anisotropy, and value-warp mechanisms with pretraining. It does not draw the turbulent component, a second smooth scale, or hyperplane jumps from the full training mixture. Its role is to test an unseen dimensionality within that construction family.

Dimension-restricted baselines receive deterministic reshapes that preserve flat element order. \rebuttalupdate{SZ3}, ZFP and HPEZ collapse leading axes into one when the tensor has more than four dimensions, retaining the last three axes. HPEZ also removes singleton axes. SPERR removes singleton axes and collapses leading axes into one depth axis when the tensor has more than three dimensions, retaining the last two axes. It codes the resulting volume in depth slabs targeting at most $2^{26}$ values, with boundary adjustments to avoid a singleton depth. These reshapes require no padding on the evaluated fields. \rebuttalupdate{TTHRESH removes} singleton axes and retains the remaining dimensions. HPEZ uses its full feature configuration, $q=4$. Pointwise tolerances are relative to the range of the original field, including when a reshaped volume is coded in slabs.

The benchmark computes BPP from the returned byte stream. For LATEC this includes the stream header, normalization extrema, gate descriptors, \rebuttalupdate{entropy-coder stage descriptors (including $\kappa$), }residual payload, and float32 escapes. Pretrained or fine-tuned model weights are supplied separately and excluded from BPP. Small dictionaries passed separately by benchmark adapters, such as shape, dtype, and checkpoint location, are also excluded.

\subsection{Learned baseline evaluation}
\label[appendix]{app:caesar}

We evaluated CAESAR because it is, to the best of our knowledge, the only other learned scientific-data compressor with a publicly available pretrained model that can be applied directly to our benchmark~\citep{li2025foundation,li2025caesar}. CAESAR provides two pretrained spatiotemporal variants. CAESAR-V combines a variational autoencoder, scale hyperprior, and super-resolution, while CAESAR-D applies conditional latent diffusion to autoencoded keyframes. Both use a PCA residual post-process that bounds global $L^2$ error rather than pointwise $L^\infty$ error directly.

Our adapter maps the last three axes to time, height, and width, merges preceding axes into channels, and uses a batch axis of one. The decoder removes temporal padding and restores the original shape. This fixed interpretation is an adapter convention, not a claim that every field's axes represent those modalities. Both variants failed to complete on five higher-dimensional fields and required substantially more BPP than SZ3 wherever they completed, under both matched-RME and matched-PSNR evaluation. \Cref{tab:caesar-results} reports the complete results.

\begin{table*}[t]
\centering
\footnotesize
\caption{CAESAR average relative BPP difference (\%; lower is better) at matched quality, relative to SZ3. Negative values denote a BPP saving. Blue, green, and orange column backgrounds denote 3-D, 4-D, and 5-D datasets, respectively. Matched-RME results were interpolated because CAESAR does not provide an RME bound. \texttt{ERR} denotes a configuration that did not complete successfully. See \Cref{app:rate-gains} for the calculation.}
\label{tab:caesar-results}
\vspace{-4pt}
\newcommand{\rotateheader}[1]{\makebox[0pt][c]{\rotatebox[origin=lb]{90}{\strut #1}}}
\begin{tblr}{
  colspec = {ll*{13}{r}},
  colsep = 1.5pt,
  row{1} = {halign=c,valign=b,belowsep=2pt,cmd=\rotateheader},
  cell{1}{1-2} = {cmd={},halign=l},
  cell{2,4}{1} = {r=2}{halign=c,valign=m},
  column{3-7} = {bg=blue!8},
  column{8-12} = {bg=green!10},
  column{13-15} = {bg=orange!12},
  hline{2,4} = {solid},
}
Metric & Method & Gray-Scott & \shortstack[l]{Euler-Multi-\\Quadrants} & \shortstack[l]{Hurricane\\ISABEL} & Miranda & S3D & \shortstack[l]{Gravity-\\Cooling} & MHD-64 & \shortstack[l]{Supernova-\\Explosion} & \shortstack[l]{RT-\\Instability} & \shortstack[l]{Convective-\\Envelope} & GyroSwin & Synthetic 5D & Gysela \\
\rotatebox{90}{RME} & CAESAR-D & 835.2 & 524.2 & 689.2 & 697.0 & 2604.1 & \texttt{ERR} & \texttt{ERR} & 3250.1 & 748.8 & 6506.0 & \texttt{ERR} & \texttt{ERR} & \texttt{ERR} \\
 & CAESAR-V & 775.6 & 436.7 & 701.5 & 669.2 & 2150.7 & \texttt{ERR} & \texttt{ERR} & 2774.1 & 715.9 & 5016.1 & \texttt{ERR} & \texttt{ERR} & \texttt{ERR} \\
\rotatebox{90}{PSNR} & CAESAR-D & 224.2 & 327.0 & 533.7 & 163.0 & 2591.6 & \texttt{ERR} & \texttt{ERR} & 251.6 & 499.7 & 196.0 & \texttt{ERR} & \texttt{ERR} & \texttt{ERR} \\
 & CAESAR-V & 239.6 & 262.2 & 494.4 & 148.4 & 1917.0 & \texttt{ERR} & \texttt{ERR} & 249.6 & 484.0 & 201.4 & \texttt{ERR} & \texttt{ERR} & \texttt{ERR} \\
\end{tblr}
\vspace{-8pt}
\end{table*}

The error targets in this benchmark are far from CAESAR's intended operating regime. CAESAR is designed for very high compression ratios, where the autoencoded latent carries the field and the PCA residual stage only refines it. At our tighter targets, the latent is nearly a fixed cost while the residual correction supplies almost all of the accuracy, so most transmitted bits belong to the PCA residual stream, and results show that CAESAR is poorly matched to our evaluation regime.

\subsection{Computing the rate gains in the tables}
\label[appendix]{app:rate-gains}

\Cref{tab:error-bpp-saving,tab:distortion-bpp-saving} summarize the rate-quality curves as average relative BPP differences with respect to SZ3. At each matched quality level, with SZ3 rate $r_{\mathrm{SZ3}}$ and compared-method rate $r_m$, we compute
\begin{equation}
  g_m = \frac{r_m}{r_{\mathrm{SZ3}}}-1.
\end{equation}
We then average $g_m$ over the matched levels. Negative values therefore mean that the method uses fewer bits per point than SZ3 at matched quality, whereas positive values mean that it uses more.

For both tables, we evaluate the compressors at logarithmically spaced RME targets from $10^{-1}$ to $10^{-6}$\rebuttalupdate{ on every field}. For the maximum-error table, the matched levels are the error bounds configured for SZ3. For compressors that are also parameterized by an error bound, we retain only bounds run by both methods and use the median BPP over repeated runs at each bound. TTHRESH has no configured RME bound, so we estimate its BPP at each SZ3 bound from its achieved-error curve by log-log interpolation of BPP against achieved range-relative maximum error; targets outside its measured range are omitted. This is why its entries are marked as estimates. We apply the same procedure to the CAESAR results in \Cref{tab:caesar-results}.

For the PSNR table, we first form each method's curve from the median $(\mathrm{PSNR},\mathrm{BPP})$ of each parameter configuration. For a dataset, we identify the PSNR interval covered by every solver with at least two valid points, sample it uniformly at 100 values, and estimate each BPP by linear interpolation in PSNR and logarithmic interpolation in BPP. The reported value is the mean of $g_m$ over these 100 common-PSNR samples. No extrapolation is performed in either table. The RME averages can therefore contain different target subsets for different methods, whereas the PSNR comparison uses one common interval per field. The latter interval depends on the methods with valid curves.

\section{Additional experimental analyses}

\subsection{Rate-distortion at matched PSNR}
\label[appendix]{app:psnr-table}

\Cref{tab:distortion-bpp-saving} is the matched-PSNR counterpart of \Cref{tab:error-bpp-saving}, computed as described in \Cref{app:rate-gains}. It measures average squared error at a given rate rather than the pointwise bound this work targets, and is reported here for completeness; the body discusses why the two views differ.

\begin{table*}[t]
\centering
\fontsize{9}{9}\selectfont
\caption{Average percentage BPP difference from SZ3 across matched PSNR levels, so negative values are rate savings and lower is better; \Cref{app:rate-gains} gives the calculation. Blue, green, and orange backgrounds mark 3-D, 4-D, and 5-D datasets. A slash marks a dataset \rebuttalupdate{without fine-tuning data, hence without a fine-tuned checkpoint}. Bold is best, underlined second-best.}
\label{tab:distortion-bpp-saving}
\vspace{-4pt}
\newcommand{\rotateheader}[1]{\makebox[0pt][c]{\rotatebox[origin=lb]{90}{\strut #1}}}
\begin{tblr}{
  colspec = {l*{13}{r}},
  colsep = 1.5pt,
  row{1} = {halign=c,valign=b,belowsep=2pt,cmd=\rotateheader},
  cell{1}{1} = {cmd={},halign=l},
  column{2-6} = {bg=blue!8},
  column{7-11} = {bg=green!10},
  column{12-14} = {bg=orange!12},
  hline{2} = {solid},
}
Method & \shortstack[l]{Euler-Multi-\\Quadrants} & Gray-Scott & \shortstack[l]{Hurricane\\ISABEL} & Miranda & S3D & \shortstack[l]{Convective-\\Envelope} & \shortstack[l]{Gravity-\\Cooling} & MHD-64 & \shortstack[l]{RT-\\Instability} & \shortstack[l]{Supernova-\\Explosion} & GyroSwin & Gysela & Synthetic 5D \\
SZ3 & 0 & 0 & \underline{0} & 0 & 0 & 0 & 0 & 0 & 0 & 0 & 0 & 0 & 0 \\
ZFP & 129.5 & 38.8 & 50.2 & 111.2 & 873.6 & 37.0 & 22.4 & 11.3 & 13.6 & 83.8 & \rebuttalupdate{23.1} & 37.5 & 2.8 \\
SPERR & \underline{-3.2} & \textbf{-33.3} & 113.8 & \underline{-47.9} & \underline{-53.7} & 73.4 & 4.4 & -9.7 & 16.5 & 115.1 & \rebuttalupdate{-8.4} & \underline{-35.9} & -1.8 \\
HPEZ & \textbf{-37.5} & -4.2 & 19.6 & -24.5 & -27.7 & 36.6 & 3.2 & 1.4 & 10.8 & 72.4 & \rebuttalupdate{7.7} & 23.1 & 4.2 \\
TTHRESH & 176.3 & \underline{-20.3} & 399.6 & \textbf{-66.5} & \textbf{-66.3} & \textbf{-4.0} & 47.3 & -11.4 & 139.1 & 43.4 & \rebuttalupdate{\textbf{-38.6}} & \textbf{-91.9} & \underline{-20.3} \\
\textbf{LATEC} & \rebuttalupdate{16.7} & \rebuttalupdate{-0.8} & \rebuttalupdate{\textbf{-13.5}} & \rebuttalupdate{-10.2} & \rebuttalupdate{5.8} & \rebuttalupdate{2.4} & \rebuttalupdate{\underline{-8.8}} & \rebuttalupdate{\underline{-17.9}} & \rebuttalupdate{\underline{-33.5}} & \rebuttalupdate{\underline{-6.8}} & \rebuttalupdate{-13.1} & \rebuttalupdate{4.7} & \rebuttalupdate{\textbf{-46.6}} \\
\shortstack[l]{\textbf{LATEC}\\(fine-tuned)} & \rebuttalupdate{10.5} & \rebuttalupdate{-16.1} & / & / & / & \rebuttalupdate{\underline{-1.8}} & \rebuttalupdate{\textbf{-18.1}} & \rebuttalupdate{\textbf{-20.7}} & \rebuttalupdate{\textbf{-37.5}} & \rebuttalupdate{\textbf{-14.6}} & \rebuttalupdate{\underline{-18.8}} & \rebuttalupdate{-5.4} & / \\
\end{tblr}
\vspace{-8pt}
\end{table*}

\subsection{Ablation studies}
\label[appendix]{app:ablations}

\begin{figure}[h]
\centering
\includegraphics[width=\linewidth]{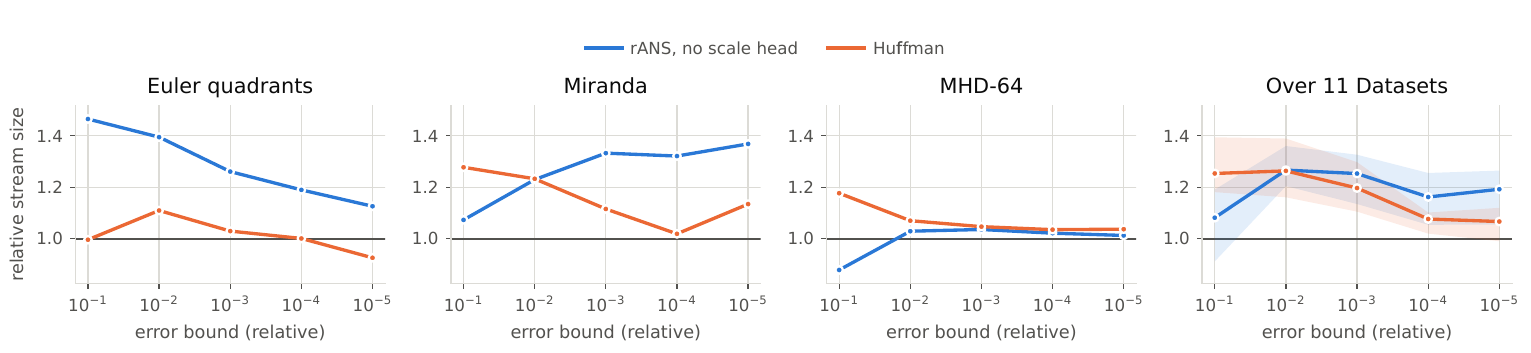}
\caption{Ablations of scale estimation and entropy coding. In the rightmost panel, each line is the mean over 11 datasets and its shaded band spans the 25th to 75th percentiles.}
\label{fig:ablation-entropy}
\end{figure}

\begin{figure}[h]
\centering
\includegraphics[width=\linewidth]{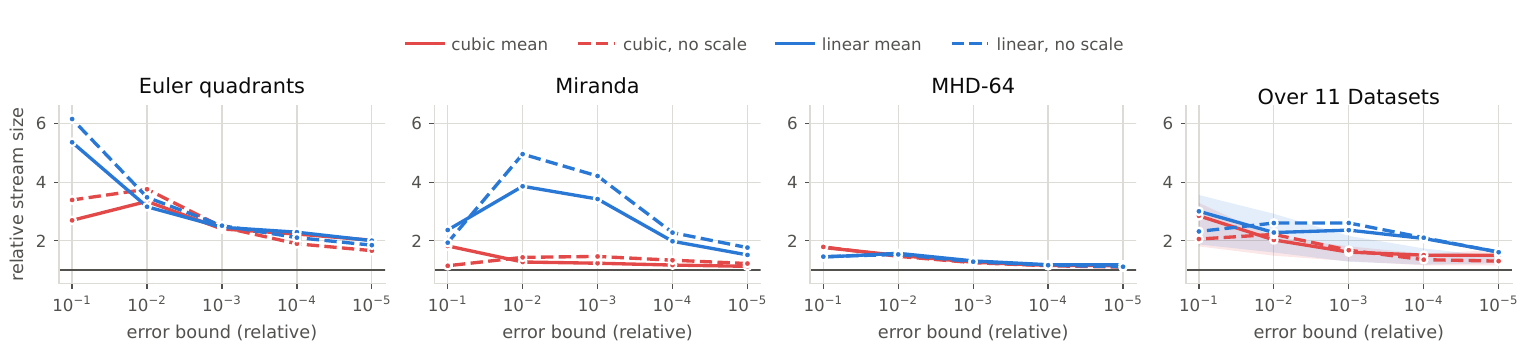}
\caption{Ablations of the prediction mean and its interaction with scale estimation. In the rightmost panel, each line is the mean over 11 datasets and its shaded band spans the 25th to 75th percentiles.}
\label{fig:ablation-predictor}
\end{figure}

We examine how the two learned outputs, the GNN mean and scale, contribute to compression. The mean provides a data-adaptive prediction that complements classical interpolation. The scale serves both as a spatially varying entropy parameter and as the pointwise signal used to gate between these predictions. \Cref{fig:ablation-entropy,fig:ablation-predictor} report component ablations on Euler-Multi-Quadrants, Miranda, and MHD-64, together with an aggregate over 11 datasets, at relative error bounds from $10^{-1}$ to $10^{-5}$. Stream sizes are normalized by the full LATEC configuration at the same bound, so a ratio above one indicates a larger stream. \Cref{fig:gate-selection} complements these rate comparisons with the fractions of points assigned to each predictor family.

\rebuttalupdate{\Cref{fig:ablation-main} is measured with the current codec on crops of at most 40M points at RME $10^{-3}$. Unlike the fixed means of \Cref{fig:ablation-predictor}, which drop the gate as well, each arm removes one component from LATEC: the mean (a classical gate over cubic and linear interpolation, with the same signaling and learned scale), the gate (GNN alone), and the scale as entropy model (\Cref{tab:ablation-entropy}). In the scale ablation, the encoder runs unchanged, so prediction, gating and reconstruction are LATEC's, and the quantized residuals of every chunk-stage are also coded under the flat Laplace scale that minimizes that stage's code length, or with a static Huffman code. Because these are crops at a single bound, the percentages differ from the full-field averages of \Cref{tab:error-bpp-saving}.}

\rebuttalupdate{The pointwise learned scale is worth 20-58\% of the bits on Hurricane, Miranda, RT-Instability and GyroSwin at RME $10^{-3}$, and 4-7\% on MHD-64 and Synthetic 5D, whose residuals are closer to stationary within a chunk-stage. On S3D, where the GNN is almost never selected and the scale head was trained on GNN residuals, the best flat scale is 3-6\% cheaper. Huffman coding, used by SZ3, is much costlier on low-rate fields because it spends at least one bit per symbol: on S3D, at 0.02 BPP, it multiplies the rate by 51.}

\begin{table}[h]
\centering
\fontsize{8}{9}\selectfont
\setlength{\tabcolsep}{4pt}
\caption{\rebuttalupdate{Entropy-model ablation on the crops of \Cref{fig:ablation-main}. LATEC's mean, gate and quantized residuals are kept, and only the coder changes: ``flat'' codes each chunk-stage under its best single Laplace scale, ``Huffman'' with a static Huffman code per chunk-stage, table included. BPP of LATEC, and BPP difference from it (\%) for each coder; lower is better.}}
\label{tab:ablation-entropy}
\rebuttalupdate{\begin{tabular}{lrrrrrr}
\toprule
 & \multicolumn{3}{c}{RME $10^{-3}$} & \multicolumn{3}{c}{RME $10^{-4}$} \\
\cmidrule(lr){2-4}\cmidrule(lr){5-7}
Field & LATEC BPP & Flat & Huffman & LATEC BPP & Flat & Huffman \\
\midrule
Hurricane & 0.320 & +58.4 & +291.0 & 0.777 & +42.8 & +117.1 \\
Miranda & 0.201 & +25.0 & +448.6 & 0.755 & +8.0 & +102.5 \\
S3D & 0.020 & -3.3 & +5046.0 & 0.111 & -6.3 & +848.9 \\
MHD-64 & 4.281 & +3.6 & +4.4 & 7.571 & +2.1 & +3.2 \\
RT-Instability & 0.614 & +55.2 & +154.3 & 1.228 & +45.2 & +80.5 \\
GyroSwin & 1.233 & +19.8 & +46.3 & 3.604 & +10.7 & +9.4 \\
Synthetic 5D & 2.708 & +7.0 & +9.1 & 5.783 & +4.6 & +5.8 \\
\bottomrule
\end{tabular}}
\end{table}

\subsubsection{Scale estimation and entropy coding}
The learned scale improves compression overall (\Cref{fig:ablation-entropy}). Across the 11-dataset aggregate, removing the scale head increases mean stream size at every bound. The field-level panels show that the gain is not uniform: the scale is consistently valuable on Euler-Multi-Quadrants and Miranda, while on MHD-64 it has little effect at tighter bounds and is detrimental at the loosest one. The aggregate trend nevertheless shows that adapting the residual distribution point by point is preferable to rANS without the learned pointwise scale. Replacing rANS with Huffman coding also increases mean stream size at every bound, confirming that the learned scale is useful when coupled to a coder that can exploit its probability estimates.

\begin{figure}[t]
\centering
\includegraphics[width=.9\linewidth]{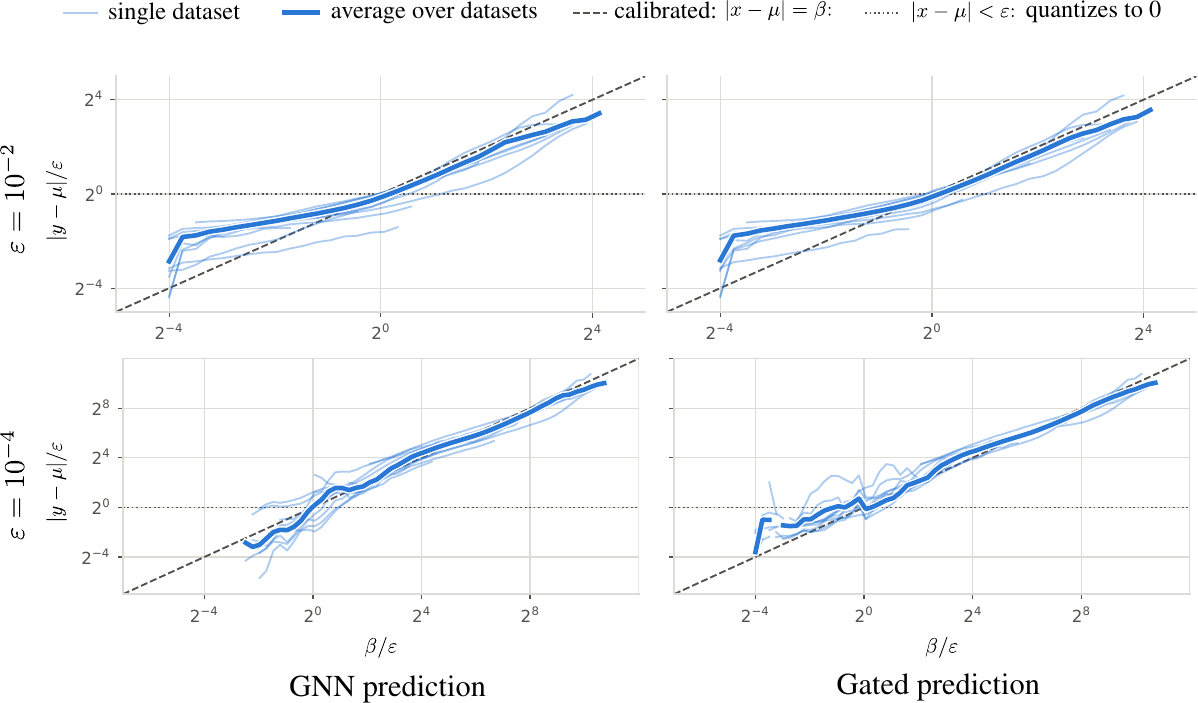}
\caption{Consistency between the predicted Laplace scale and empirical residual magnitude for the raw GNN mean and the gated mean at relative error bounds $10^{-2}$ and $10^{-4}$. At each predicted scale, a curve reports the mean absolute residual of the points assigned that scale. Each light curve corresponds to one dataset, and the dark curve averages across datasets. The dashed diagonal shows the moment relation $\mathbb{E}|y_p-\widehat\mu_p|=\widehat\beta_p$ implied by a zero-centered Laplace model. The dotted horizontal line is the boundary $|y_p-\widehat\mu_p|=\varepsilon_i$, below which a residual quantizes to the zero index.}
\label{fig:scale-calibration}
\end{figure}

The scale $\widehat\beta_p$ is not trained to estimate the individual magnitude $|y_p-\widehat\mu_p|$. It parameterizes the conditional Laplace distribution from which the entropy model assigns residual probabilities. This distribution nevertheless implies the group-level moment $\mathbb{E}|y_p-\widehat\mu_p|=\widehat\beta_p$. Equivalently, for residuals sharing one scale, their Laplace negative log-likelihood is minimized when that scale equals their mean absolute magnitude. We therefore group points by predicted scale in \Cref{fig:scale-calibration} and compare this empirical mean with the scale itself. The comparison tests one consequence of the assumed distribution, rather than pointwise residual accuracy or full distributional calibration. It is relevant to coding because a scale that is systematically too small or too large assigns an unnecessarily narrow or broad residual distribution and increases expected code length.

Both the raw GNN and gated scales satisfy this moment relation approximately over most of the represented range. At the tighter $10^{-4}$ bound, the dataset-level curves broadly follow the diagonal across several scale levels. At the looser $10^{-2}$ bound, they flatten at small scales and lie mostly below the boundary marking the zero-index region. Scale mismatch is less consequential there because deviations that do not move a residual across a quantization boundary leave its coded symbol unchanged. Gating changes the low-scale behavior more than the high-scale behavior, yet preserves the overall ordering between scale and empirical residual magnitude. The scale can therefore support both predictor selection and entropy coding without estimating each residual or matching the full residual distribution exactly.

\subsubsection{Prediction and scale must be considered together}
The GNN mean is important because no fixed interpolation mean matches the full adaptive predictor across datasets (\Cref{fig:ablation-predictor}). Replacing it and its gate with either cubic or linear interpolation increases the mean stream size at every tested bound. Cubic interpolation is consistently the stronger fixed alternative in the 11-dataset aggregate, yet it remains more costly than the complete predictor. Retaining the learned scale usually reduces the cost of both fixed means, although the ordering can reverse at individual operating points. The gain therefore comes from the combination of a learned mean, classical fallbacks, and a scale model matched to the resulting residuals, rather than from one universally best prediction rule.

\begin{figure}[t]
\centering
\includegraphics[width=\linewidth]{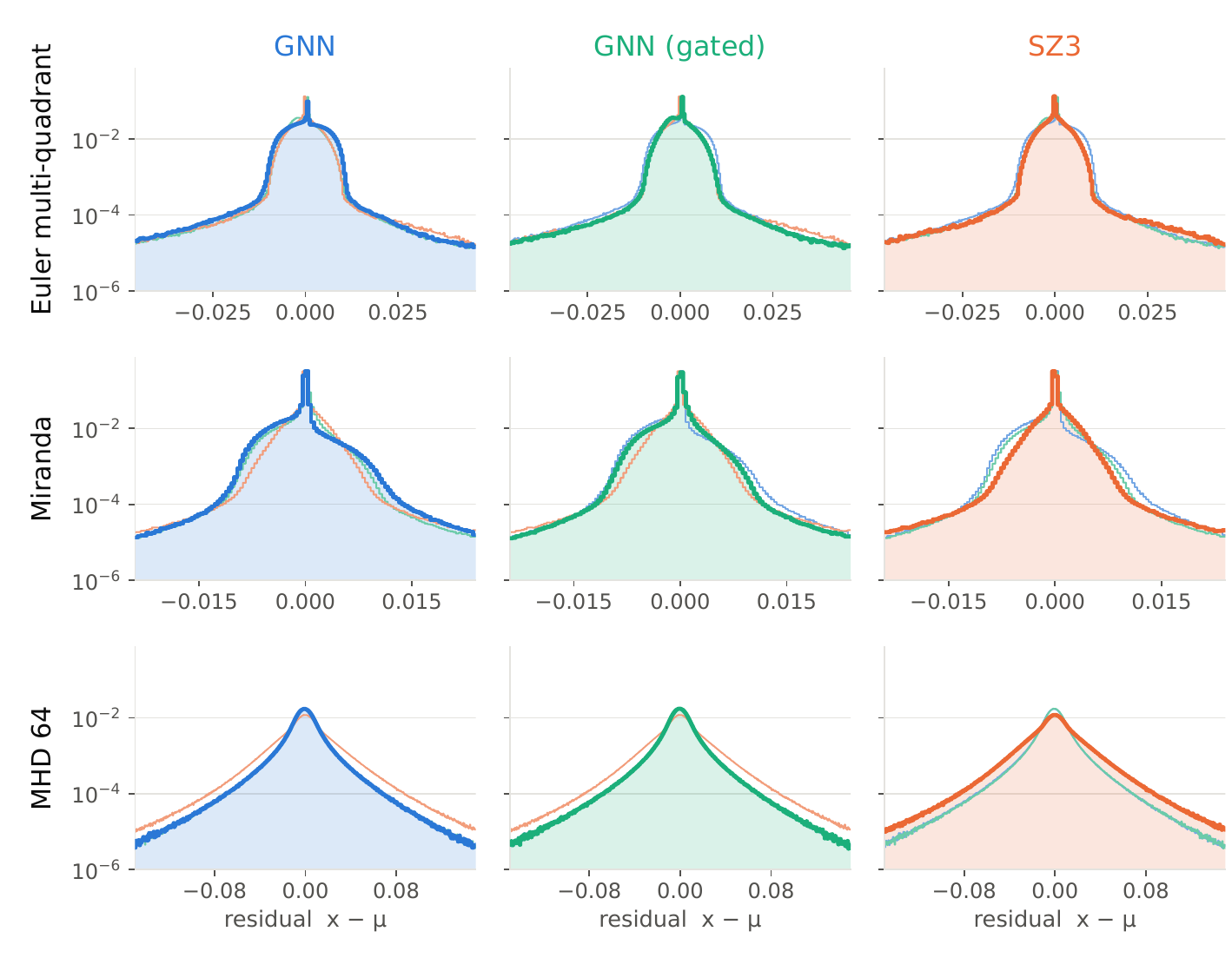}
\caption{Distributions of prediction residuals $y_p-\mu_p$ at $\varepsilon=0.01$ for Euler-Multi-Quadrants, Miranda, and MHD-64. Columns compare the raw GNN, GNN with gating, and SZ3.}
\label{fig:residual-distribution}
\end{figure}

The residual distributions in \Cref{fig:residual-distribution} show how prediction errors differ across the same three fields. At a relative error bound of $10^{-2}$, all predictors concentrate residuals near zero, but their tails and structures vary by field.

\begin{figure}[h]
\centering
\includegraphics[width=\linewidth]{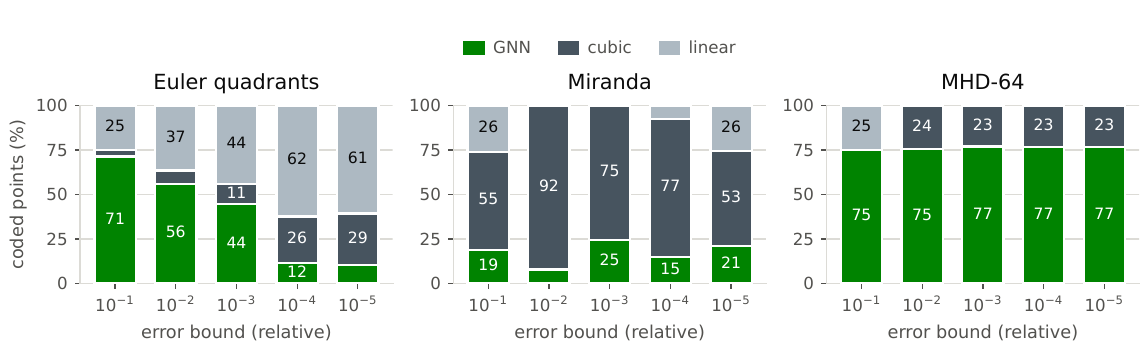}
\caption{Fraction of coded points assigned by the full gate to the GNN, cubic interpolation, and linear interpolation.}
\label{fig:gate-selection}
\end{figure}

\subsubsection{Adaptive predictor selection}
The learned scale also enables fine-grained selection of the prediction mean at low signaling cost (\Cref{fig:gate-selection}). Although the encoder transmits only one threshold rule per chunk-stage, the learned scale $\beta_p$ varies across points, so applying that rule produces a pointwise gate without transmitting pointwise flags. MHD-64 assigns about three quarters of its points to the GNN throughout the tested range, with nearly all remaining points assigned to linear interpolation. Miranda instead relies mainly on cubic interpolation, while Euler-Multi-Quadrants shifts from 71\% GNN selection at $10^{-1}$ toward predominantly linear interpolation at the tightest bounds. These distinct selections show that the GNN mean is highly useful but not uniformly optimal. The scale-guided gate retains it where it is competitive and substitutes a classical mean elsewhere, combining local adaptivity with one compact gate descriptor, including its scale shift $s$, per chunk-stage.

\begin{figure}[h]
\centering
\includegraphics[width=\linewidth]{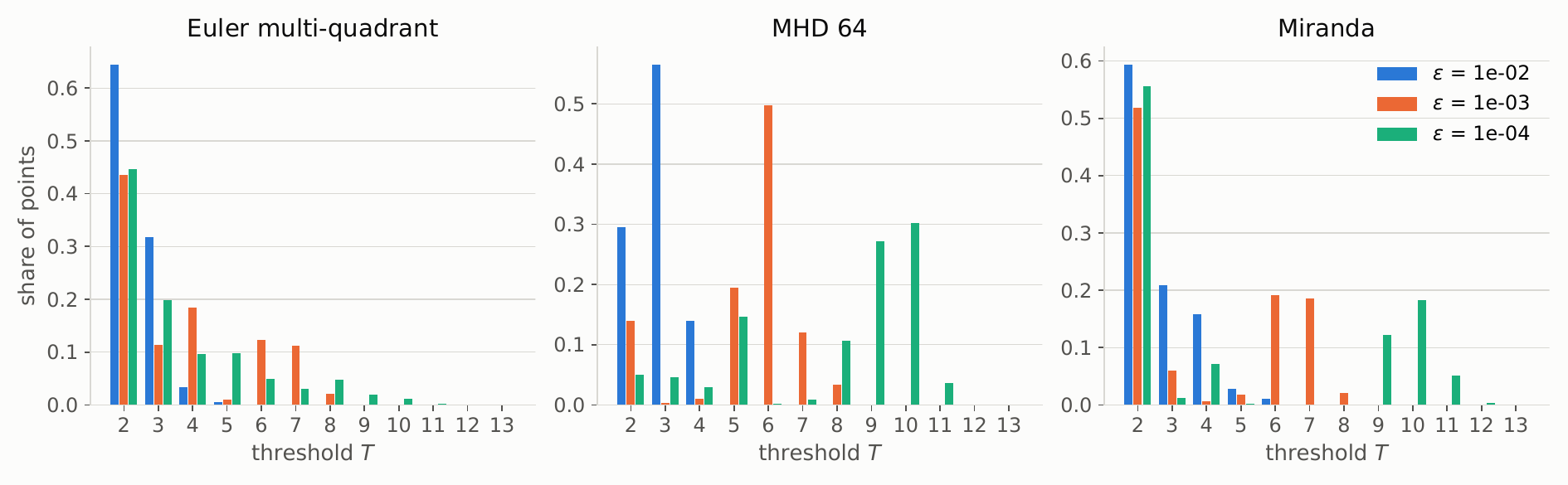}
\includegraphics[width=\linewidth]{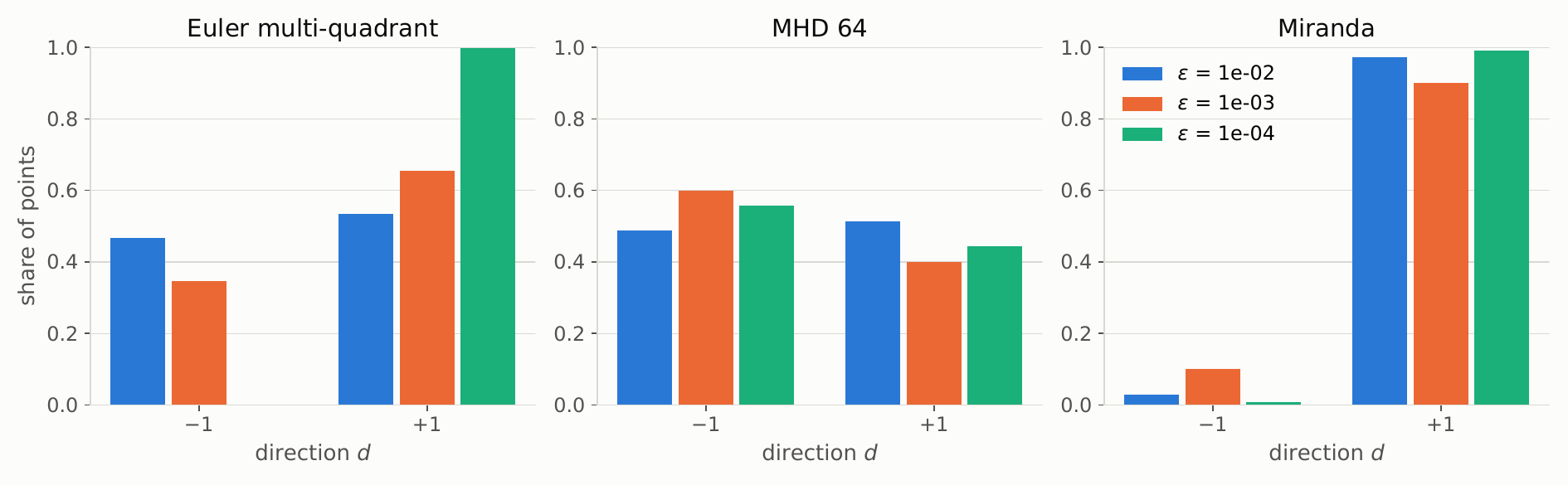}
\includegraphics[width=\linewidth]{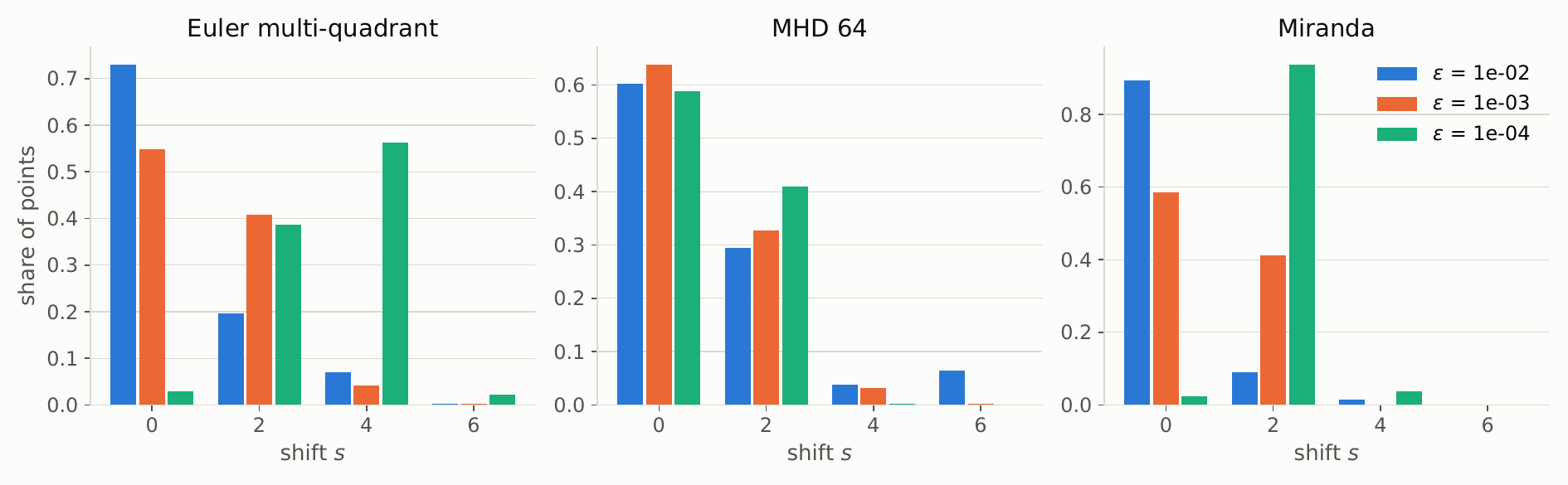}
\caption{Distribution of the selected gate parameters of \Cref{eq:gate} over gated points (top to bottom: threshold exponent $T$, direction $d$, scale shift $s$), for Euler-Multi-Quadrants, MHD-64, and Miranda at $\varepsilon\in\{10^{-2},10^{-3},10^{-4}\}$.}
\label{fig:gate-params}
\end{figure}

\Cref{fig:gate-params} breaks the selected gates down by their remaining parameters: the threshold exponent $T$, the direction $d$, and the scale shift $s$. The selected threshold moves toward larger exponents as the bound tightens, most visibly on MHD-64, since the scales $\beta_p$ grow relative to $\varepsilon_i$ when the bound tightens. Miranda almost always selects $d=+1$, which replaces the GNN mean on the low-scale points, whereas MHD-64 uses both directions in similar proportions. The shift is mostly $s\in\{0,2\}$, with $s=4$ appearing on Euler-Multi-Quadrants at $10^{-4}$ and $s=6$ remaining rare, so the scale correction is usually moderate.

\rebuttalupdate{\subsubsection{Model width}
\label[appendix]{app:model-width}
\Cref{tab:ablation-width} compares the shipped width of 64 with retrained widths of 16 and 32, on the 12 benchmark fields other than Gysela. Width 64 is best on 11 of the 12 fields. On the remaining one, Gravity-Cooling, the three widths lie within 3.8 points. Width 32 uses on average 4.5\% more bits than width 64, for a modest speedup on the H100 setup of \Cref{app:runtime-memory}: 1.01$\times$ in encoding and 1.27$\times$ in decoding in 3-D, and 1.59$\times$ and 1.47$\times$ in 5-D. Halving the width again to 16 does not speed up encoding any further (1.06$\times$ and 1.57$\times$ over width 64 in 3-D and 5-D), and speeds up decoding only to 1.42$\times$ and 1.82$\times$. The speedup is small because runtime is dominated by per-point work that does not depend on the width, such as building each point's causal context, rather than by the network's matrix products. Width 16 is far too small: it uses more bits than SZ3 on four fields, and 11 times as many on S3D.}

\begin{table*}[t]
\centering
\fontsize{9}{9}\selectfont
\caption{\rebuttalupdate{Predictor width: average percentage BPP difference from SZ3 across matched maximum-error levels, as in \Cref{tab:error-bpp-saving}, for hidden widths 16, 32, and 64 (shipped), with their parameter counts. Negative values are rate savings and lower is better. Blue, green, and orange backgrounds mark 3-D, 4-D, and 5-D datasets. Bold is best, underlined second-best.}}
\label{tab:ablation-width}
\vspace{-4pt}
\newcommand{\rotateheader}[1]{\makebox[0pt][c]{\rotatebox[origin=lb]{90}{\strut #1}}}
\rebuttalupdate{\begin{tblr}{
  colspec = {l*{12}{r}},
  colsep = 3.5pt,
  row{1} = {halign=c,valign=b,belowsep=2pt,cmd=\rotateheader},
  cell{1}{1} = {cmd={},halign=l},
  column{2-6} = {bg=blue!8},
  column{7-11} = {bg=green!10},
  column{12-13} = {bg=orange!12},
  hline{2} = {solid},
}
Width & \shortstack[l]{Euler-Multi-\\Quadrants} & Gray-Scott & \shortstack[l]{Hurricane\\ISABEL} & Miranda & S3D & \shortstack[l]{Convective-\\Envelope} & \shortstack[l]{Gravity-\\Cooling} & MHD-64 & \shortstack[l]{RT-\\Instability} & \shortstack[l]{Supernova-\\Explosion} & GyroSwin & Synthetic 5D \\
16 (6.9k) & 21.2 & -1.4 & -6.1 & 20.4 & 1032.1 & 26.7 & \textbf{-22.3} & -11.1 & -1.8 & -15.8 & -14.0 & -27.4 \\
32 (26k) & \underline{18.3} & \underline{-7.2} & \underline{-12.8} & \underline{-13.1} & \underline{-23.3} & \underline{-4.8} & \underline{-19.7} & \underline{-18.2} & \underline{-36.1} & \underline{-17.5} & \underline{-19.0} & \underline{-39.9} \\
64 (103k) & \textbf{13.1} & \textbf{-8.6} & \textbf{-16.5} & \textbf{-15.8} & \textbf{-25.7} & \textbf{-7.5} & -18.5 & \textbf{-21.2} & \textbf{-38.1} & \textbf{-17.9} & \textbf{-27.4} & \textbf{-48.6} \\
\end{tblr}}
\vspace{-8pt}
\end{table*}

\begin{figure}[h]
\centering
\includegraphics[width=.85\linewidth]{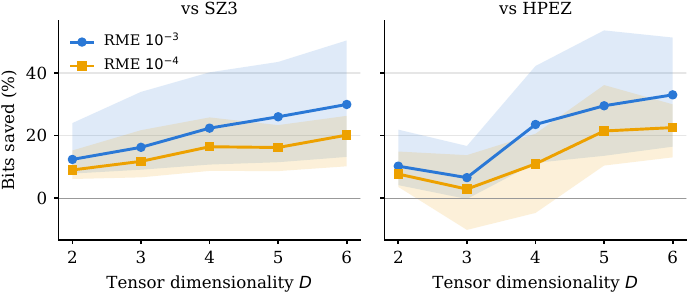}
\caption{\rebuttalupdate{Bits saved by LATEC against SZ3 (left) and HPEZ (right) on synthetic fields of 4-5M points from the pretraining generator, as a function of the number of axes $D$. Lines are means over five seeds, and shaded bands span the minimum and maximum over seeds.}}
\label{fig:dim-gain}
\end{figure}

\rebuttalupdate{\subsection{Scaling with dimensionality}
\label[appendix]{app:dimensionality}
No component of LATEC depends on the number of axes $D$, so the same pretrained weights apply to any $D$. To measure how the gain varies with $D$ alone, we compress synthetic fields from the pretraining generator with fixed generator settings for $D=2$ to $6$, with 4-5M points each ($2048^2$, $256\times128^2$, $64\times40^3$, $32\times20^4$, and $16\times12^5$), five seeds per $D$, at RME $10^{-3}$ and $10^{-4}$. The fields are in distribution except for $D$: pretraining uses 2-D and 4-D fields only, so $D=3$, 5, and 6 are unseen. The saving grows with $D$ (\Cref{fig:dim-gain}): against SZ3, from 12.4\% at $D=2$ to 29.9\% at $D=6$ at RME $10^{-3}$, and from 9.0\% to 20.2\% at $10^{-4}$. Against HPEZ on the same fields, it grows from 3-10\% at $D=2$ and $D=3$ to 23-33\% at $D=6$. The seed-to-seed spread is wide, from 13\% to 50\% against SZ3 at $D=6$ and $10^{-3}$.}

\rebuttalupdate{We also compress a real 6-D tensor, obtained by stacking $T=8$ or 16 consecutive GyroSwin timesteps, from $t=100$, along a new leading axis (\Cref{tab:real-6d}). Each timestep is cropped from 85 to 21 points along its fourth axis, giving 22M and 44M points. SZ3 and HPEZ support at most four axes, so they receive the stack with its leading axes merged, $256T\times16\times21\times32$, and the time axis is folded into the first spatial axis. Native 6-D LATEC uses 32.7-32.8\% fewer bits than SZ3 at RME $10^{-3}$ and 14.4-16.1\% fewer at $10^{-4}$, and 38\% and 18\% fewer than HPEZ. Coding the stack natively in 6-D is only 0.4-3.5\% better than coding each timestep separately in 5-D, so the time axis adds little on this dataset.}

\begin{table}[h]
\centering
\fontsize{8}{9}\selectfont
\setlength{\tabcolsep}{4pt}
\caption{\rebuttalupdate{Real 6-D data: $T$ stacked GyroSwin timesteps of shape $32\times8\times16\times21\times32$. BPP difference from SZ3 on the stack (\%) for HPEZ, LATEC coding each timestep separately in 5-D, and LATEC coding the stack natively in 6-D. SZ3 and HPEZ receive the stack reshaped to $256T\times16\times21\times32$. Lower is better, \textbf{bold} is best.}}
\label{tab:real-6d}
\rebuttalupdate{\begin{tabular}{rrrrr}
\toprule[1pt]
 & & & LATEC & LATEC \\
$T$ & RME & HPEZ & per-step & 6-D \\
\midrule[1pt]
8 & $10^{-3}$ & 8.0 & -32.4 & \textbf{-32.7} \\
 & $10^{-4}$ & 4.5 & -13.0 & \textbf{-14.4} \\
\midrule
16 & $10^{-3}$ & 8.1 & -32.4 & \textbf{-32.8} \\
 & $10^{-4}$ & 2.8 & -13.0 & \textbf{-16.1} \\
\bottomrule[1pt]
\end{tabular}}
\end{table}

\subsection{Runtime and memory measurements}
\label[appendix]{app:runtime-memory}

\begin{figure}[t]
\centering
\includegraphics[width=\linewidth]{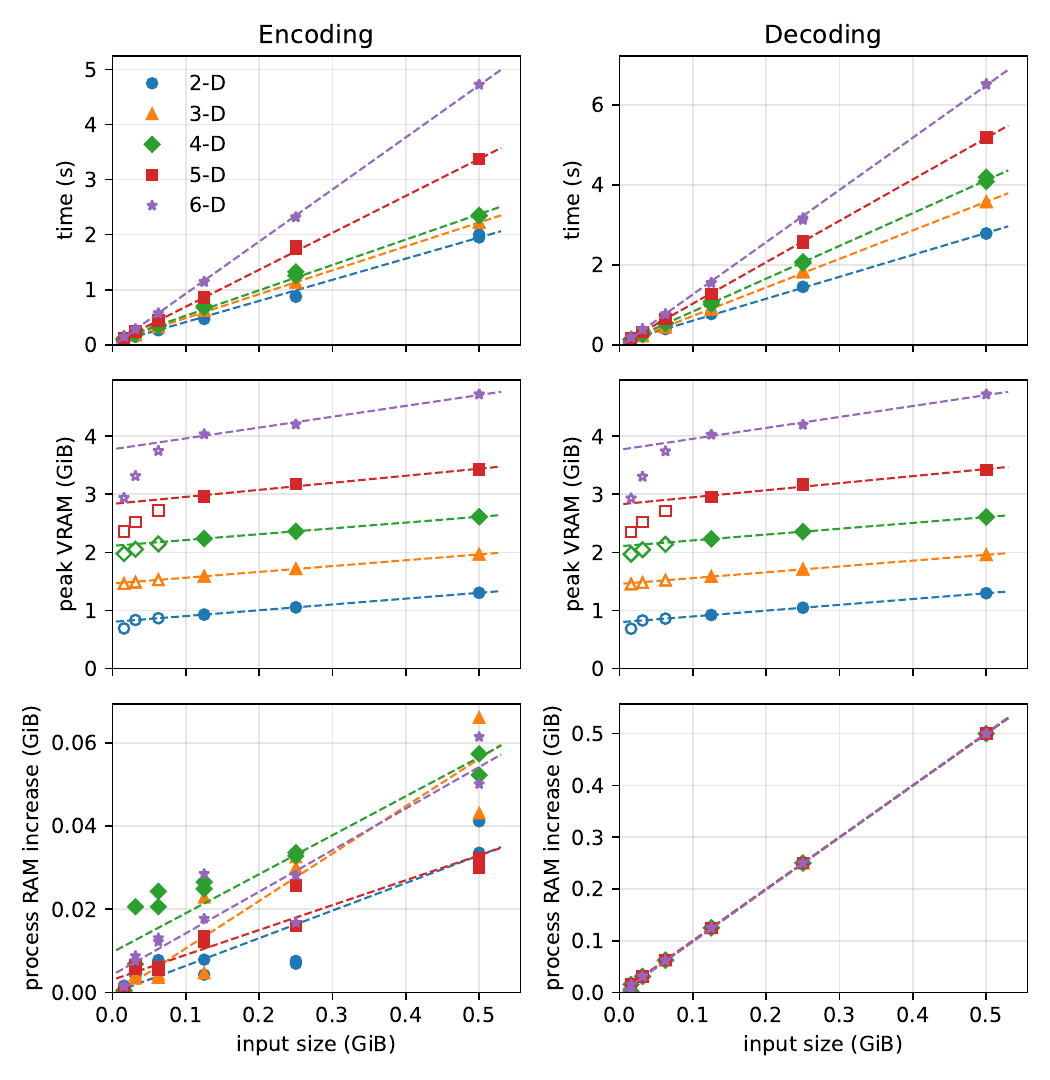}
\caption{Runtime and memory scaling of LATEC from 2-D to 6-D, encoding on the left and decoding on the right\rebuttalupdate{, measured on an H100 on synthetic fields that SZ3 compresses to about 2 bits per point, at a relative error bound of $10^{-4}$. From top to bottom: wall time, peak GPU memory, and increase of process RAM during the call. Markers show measurements and dashed lines show affine fits as a function of input size. Hollow markers are excluded from the GPU-memory fit. The comparison with the baselines is in \Cref{fig:runtime-memory}.}}
\label{fig:runtime-memory-regressions}
\end{figure}

\rebuttalupdate{All runtimes were measured on one NVIDIA H100 GPU and an AMD EPYC 9554 CPU at 3.1\,GHz. LATEC was timed with the current implementation on synthetic fields, which SZ3 compresses to about 2 bits per point, at a relative error bound of $10^{-4}$. Each dimensionality from 2-D to 6-D is measured at six sizes between 16 and 512\,MiB, on two fields per size, each in a fresh process. Time is fitted as a fixed cost plus size over throughput with a Huber regression. Process RAM is measured as its increase during the codec call, which excludes the process baseline and the input array, and is fitted by least squares on all sizes. Peak GPU memory is fitted by least squares on the three largest sizes, since the smaller ones sit below the asymptotic line. SZ3 (through pysz), SPERR and TTHRESH were run with 8 OpenMP threads on the same CPU, at 512\,MiB on three fields per dimensionality from the same generator, at the same error bound. Their peak process RAM is read from the kernel's resident-memory high-water mark, reset before each call. SZ3 and SPERR are measured only up to the ranks they support, 4-D and 3-D.} \Cref{fig:runtime-memory} compares encoding and decoding throughput and memory scaling across dimensionalities. \rebuttalupdate{\Cref{fig:runtime-memory-regressions} shows LATEC's encoding time, decoding time, peak process RAM, and peak GPU memory as functions of input size from 2-D to 6-D, together with their affine fits, and \Cref{tab:latec-regressions} lists the fitted slopes and intercepts.} The time fits provide throughput and fixed runtime overheads, while the memory fits provide memory growth per unit of input and fixed memory overheads. The fixed overhead gives LATEC a larger relative footprint on the smallest tensors, but these already fit readily in host memory; scaling matters for the tensors of up to 4\,GiB that we compress in practice. \rebuttalupdate{Encoding throughput decreases slowly from 267\,MiB/s in 2-D to 222\,MiB/s in 4-D, then to 153 and 108\,MiB/s in 5-D and 6-D, while decoding slows from 187\,MiB/s in 2-D to 78\,MiB/s in 6-D. GPU memory grows by 1.0\,MiB per MiB up to 4-D, and by 1.2 and 1.9 in 5-D and 6-D. Beyond the process baseline and the input, the codec itself adds little process RAM: decoding grows it by exactly the size of the reconstruction, and encoding by less than 0.15\,MiB per MiB of input.}

\begin{table}[t]
\centering
\caption{\rebuttalupdate{Affine fits of LATEC's runtime and memory against input size, from the measurements of \Cref{fig:runtime-memory-regressions}. Parenthesized values are fitted intercepts, in seconds for throughput and MiB for GPU memory. RAM increase is the growth of process RAM during the codec call, whose intercepts are below 20\,MiB.}}
\label{tab:latec-regressions}
\begin{tabular}{lrrrr}
\toprule
 & Encoding & Decoding & RAM increase & Peak GPU memory \\
 & MiB/s (+s) & MiB/s (+s) & enc.\ / dec., MiB/MiB & MiB/MiB (+MiB) \\
\midrule
2-D & 267 (+0.03) & 187 (+0.05) & 0.07 / 1.00 & 1.00 (+824) \\
3-D & 236 (+0.05) & 143 (+0.01) & 0.11 / 1.00 & 1.00 (+1499) \\
4-D & 222 (+0.07) & 125 (+0.01) & 0.09 / 1.01 & 1.00 (+2163) \\
5-D & 153 (+0.02) & 99 (+0.00) & 0.06 / 1.00 & 1.21 (+2902) \\
6-D & 108 (+0.00) & 78 (+0.00) & 0.10 / 1.01 & 1.87 (+3866) \\
\bottomrule
\end{tabular}
\end{table}

\subsection{Predicted fields, scales, and residuals}
\label[appendix]{app:pred-fields}

\begin{figure}[p]
\centering
\setlength{\tabcolsep}{1pt}
\renewcommand{\arraystretch}{0.6}
\newcommand{\pfim}[1]{$\vcenter{\hbox{\includegraphics[width=.21\linewidth]{#1}}}$}
\newcommand{\pfrow}[2]{%
  $\vcenter{\hbox{\rotatebox{90}{\small #2}}}$ &
  \pfim{figs/pred_fields/grayscott_#1.png} &
  \pfim{figs/pred_fields/mhd64_#1.png} &
  \pfim{figs/pred_fields/rti_#1.png} &
  \pfim{figs/pred_fields/s3d_#1.png} \\ \noalign{\vskip 2pt}}
\begin{tabular}{c@{\hskip 2pt}cccc}
& \small Gray-Scott & \small MHD-64 & \small RT-Instability & \small S3D \\ \noalign{\vskip 2pt}
\pfrow{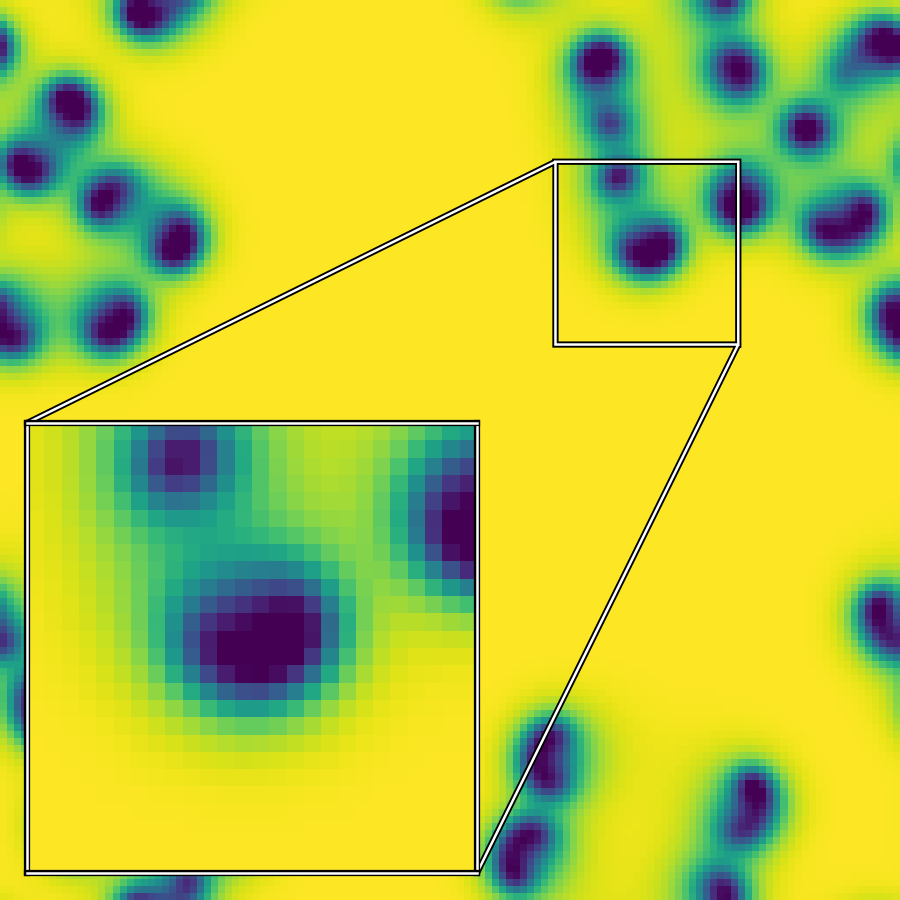}{Truth}
\pfrow{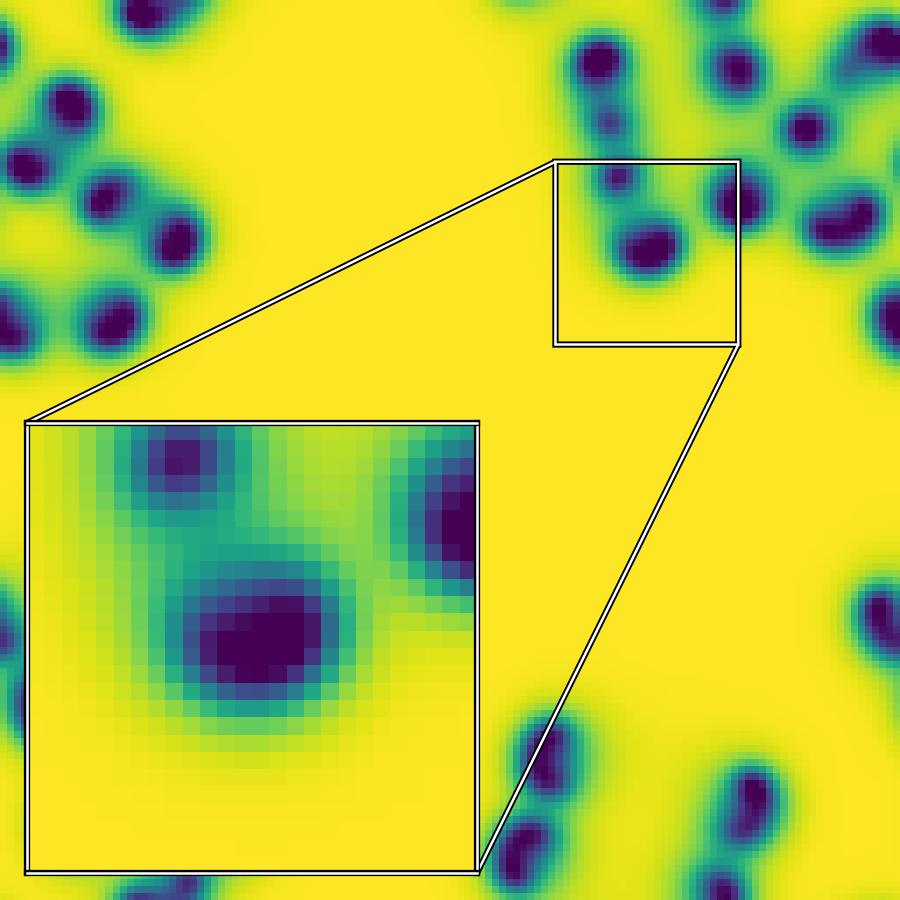}{GNN mean}
\pfrow{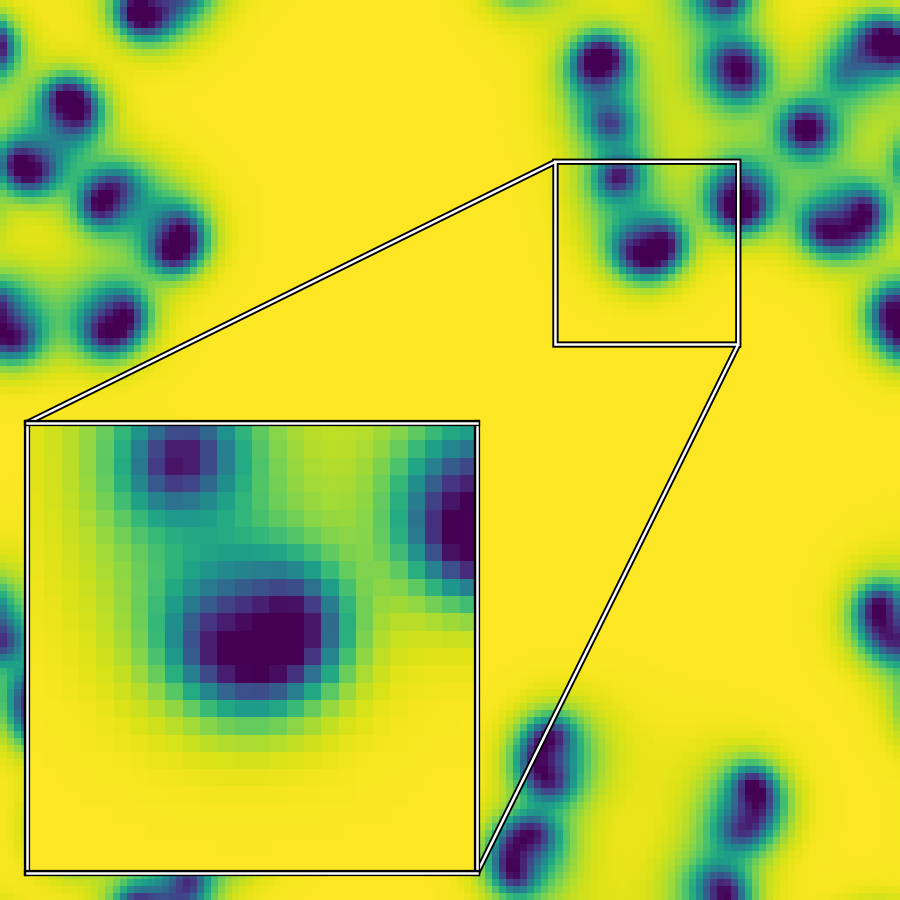}{Gated mean}
\pfrow{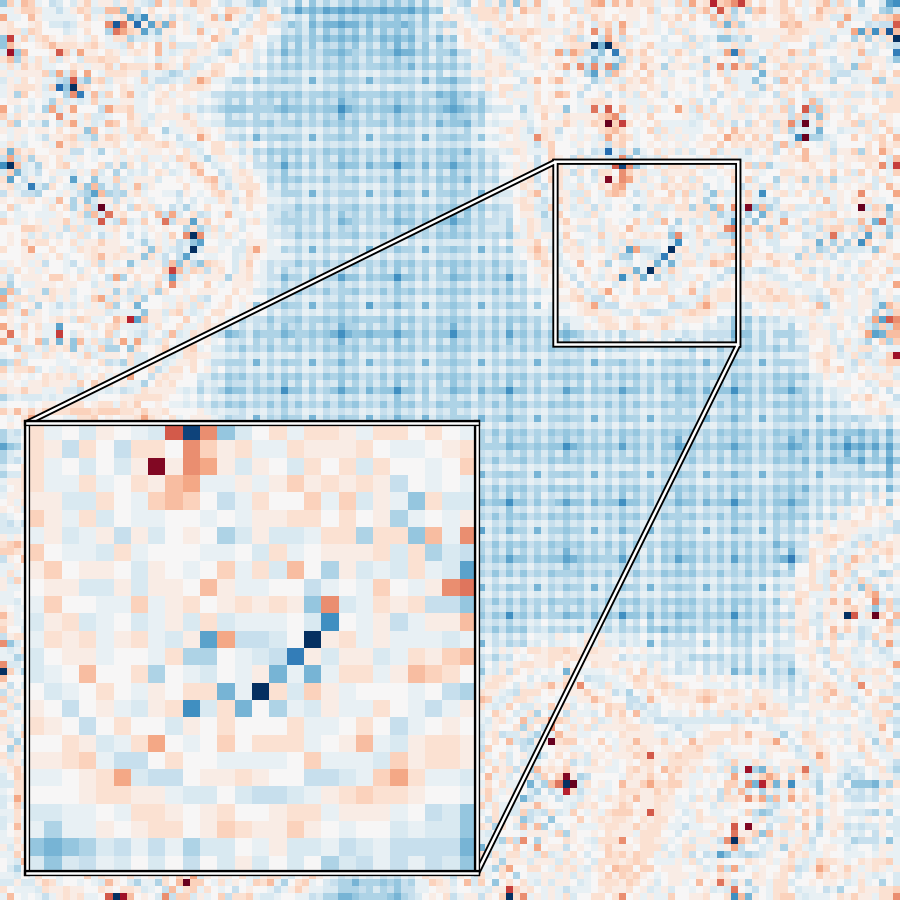}{GNN quant. resid.}
\pfrow{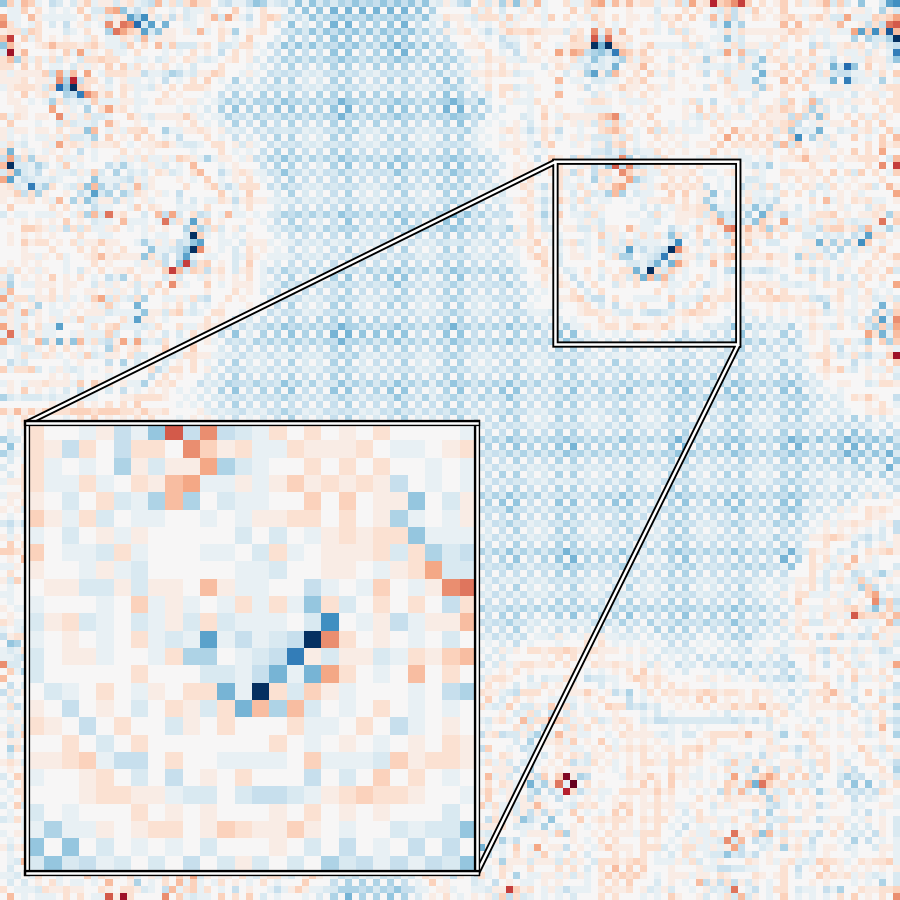}{Gated quant. resid.}
\pfrow{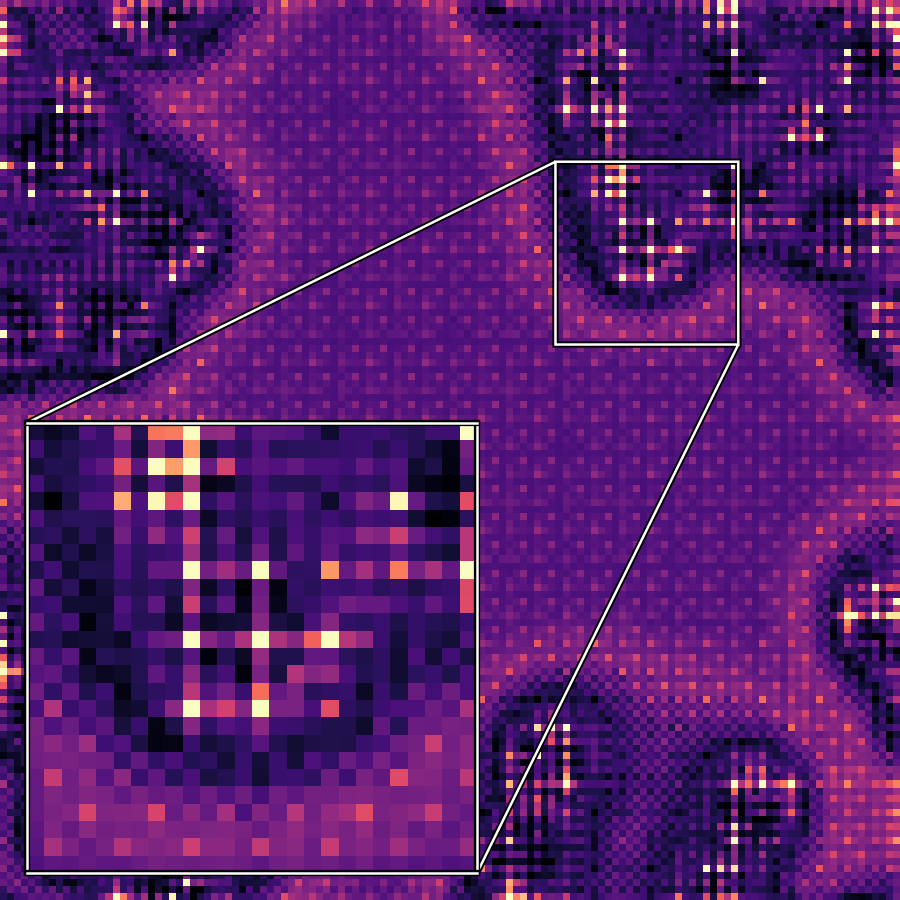}{GNN scale}
\pfrow{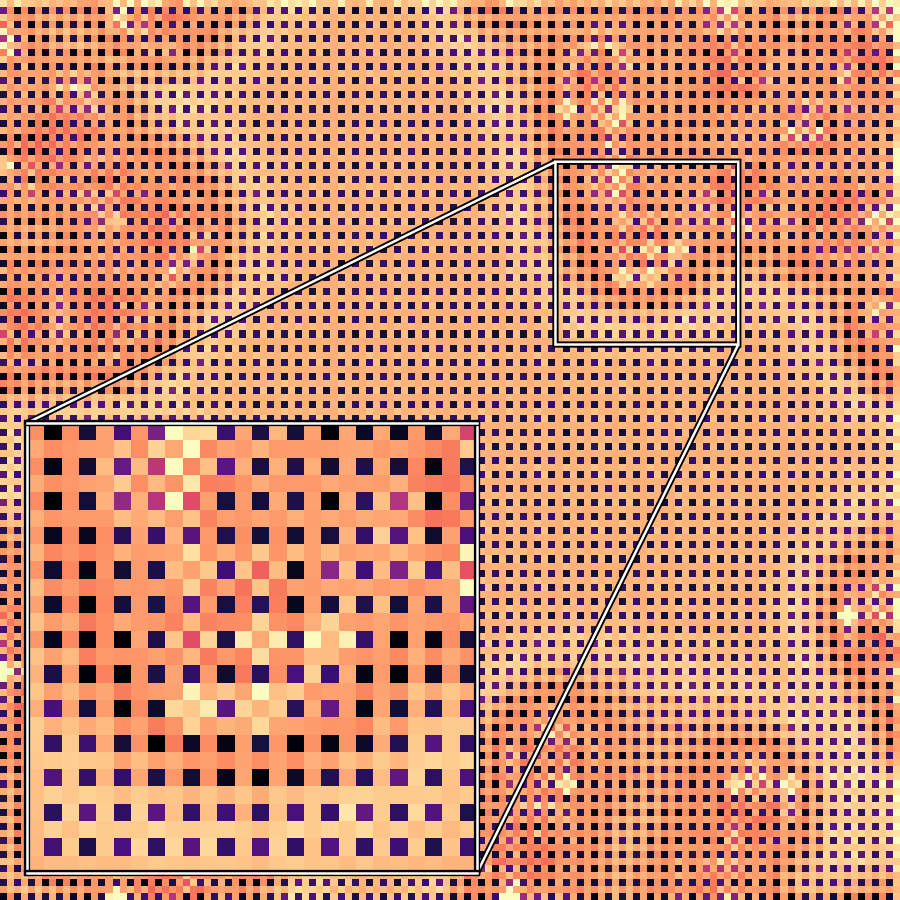}{Gated scale}
\end{tabular}
\vspace{-6pt}
\caption{Predictor outputs on a 2-D slice of four benchmark fields at relative error bound $10^{-4}$, for the raw GNN head and for the gated predictor.}
\label{fig:pred-fields}
\end{figure}

\Cref{fig:pred-fields} visualizes what the predictor produces on a 2-D slice of four benchmark fields at relative error bound $10^{-4}$. Each row shows one quantity: the ground-truth field, the mean, the quantized residual normalized by the scale, and the scale itself, the last three given both for the raw GNN head and for the gated predictor the codec actually uses. Means and residuals use a diverging blue-to-red map from negative to positive, and the scale a logarithmic map from dark to bright. Panels are scaled independently, except the two residual rows, which share one color scale per dataset so that the GNN and gated residuals can be compared directly.

The predicted mean reproduces the large-scale structure of each field, while the predicted scale concentrates on the interfaces, filaments, and shocks where the interpolation is least reliable, and stays near zero in smooth regions. The normalized residuals are correspondingly largest where the scale is largest, which is the behavior the entropy coder relies on: a spatially varying scale that tracks residual magnitude is what turns the residual field into a cheap stream. The GNN and gated rows agree on MHD-64 and RT-Instability, where the gate selects the GNN almost everywhere, and differ on Gray-Scott and S3D, where interpolation is selected for most points.

\end{document}